\documentclass[10pt, journal, compsoc]{IEEEtran}
\usepackage{amsmath, amsfonts}
\usepackage{algorithmic}
\usepackage{algorithm}
\usepackage{array}
\usepackage[caption=false, font=normalsize, labelfont=sf, textfont=sf]{subfig}
\usepackage{textcomp}
\usepackage{stfloats}
\usepackage{url}

\usepackage[edges]{forest}
\usepackage{tikz}

\usepackage{verbatim}
\usepackage{graphicx}
\usepackage{colortbl}
\usepackage{cite}

\usepackage{changepage}
\usepackage{booktabs}       
\usepackage{amsfonts}       
\usepackage{nicefrac}       
\usepackage{microtype}      
\usepackage{xcolor}         
\usepackage{adjustbox}
\usepackage{multirow}
\usepackage{makecell}
\usepackage{epsfig, epstopdf}
\usepackage{bbding}
\usepackage{wrapfig, blindtext} 
\usepackage{bm} 
\usepackage{pdfpages}

\usepackage{multirow}
\usepackage{hyperref}
\usepackage{arydshln} 
\usepackage{xspace}
\newcommand{\ie}{{\emph{i.e.}}, \xspace}

\newcommand{\etal}{{\emph{et al.}}}

\tikzset{
basic/.style = {draw, font=\footnotesize, rectangle},
root/.style = {basic, font=\bfseries\footnotesize,  text width=2.0cm, rounded corners=2pt, semithick, align=center, fill=yellow!10},
parent/.style = {basic, text width=3.3cm, rounded corners=1.5pt, thin, align=left, fill=blue!3},
child/.style = {basic, text width=11.3cm, rounded corners=1.5pt, thin, align=left, fill=gray!5},
}

\begin{document}

\title{Low-bit Model Quantization \\ for Deep Neural Networks: A Survey}

\author{
    Kai Liu$^{\dagger}$,
    Qian Zheng$^{\dagger}$,
    Kaiwen Tao$^{\dagger}$,
    Zhiteng Li,
    Haotong Qin,
    Wenbo Li,
    Yong Guo, 
    Xianglong Liu, \\
    Linghe Kong$^{*}$,
    Guihai Chen, \IEEEmembership{Fellow,~IEEE},
    Yulun Zhang$^{*}$,
    Xiaokang Yang, \IEEEmembership{Fellow,~IEEE}

    \IEEEcompsocitemizethanks{
        \IEEEcompsocthanksitem $^{\dagger}$ denotes equal contribution. 
        
        \IEEEcompsocthanksitem Kai Liu, Qian Zheng, Kaiwen Tao, Zhiteng Li, Linghe Kong, Guihai Chen, Yulun Zhang, and Xiaokang Yang are with the School of Computer Science, Shanghai Jiao Tong University, Shanghai, China. Email: 
        \href{mailto:normal.kliu@gmail.com}{normal.kliu@gmail.com},
        \href{mailto:xiaozheng2023@sjtu.edu.cn}{xiaozheng2023@sjtu.edu.cn},
        \href{mailto:miukiii19@sjtu.edu.cn}{miukiii19@sjtu.edu.cn},
        \href{mailto:ieeezhitengli@gmail.com}{ieeezhitengli@gmail.com},
        \href{mailto:linghe.kong@sjtu.edu.cn}{linghe.kong@sjtu.edu.cn},
        \href{mailto:gchen@cs.sjtu.edu.cn}{gchen@cs.sjtu.edu.cn},
        \href{mailto:yulun100@gmail.com}{yulun100@gmail.com},
        \href{mailto:xkyang@sjtu.edu.cn}{xkyang@sjtu.edu.cn}.
        
        \IEEEcompsocthanksitem Haotong Qin is with ETH Z\"{u}rich, Switzerland. (Email: \href{mailto:qinhaotong@gmail.com}{qinhaotong@gmail.com}). 
        \IEEEcompsocthanksitem Wenbo Li is with Huawei Noah's Ark Lab, China. (Email: \href{mailto:wenboli@cse.cuhk.edu.hk}{wenboli@cse.cuhk.edu.hk}). Yong Guo is with Huawei Consumer Business Group, China. (E-mail: \href{mailto:guoyongcs@gmail.com}{guoyongcs@gmail.com}).

        \IEEEcompsocthanksitem Xianglong Liu is with Beihang University, China. (Email: \href{mailto:xlliu@buaa.edu.cn}{xlliu@buaa.edu.cn}).
        \IEEEcompsocthanksitem Yulun Zhang and Linghe Kong are the corresponding authors. 
    }

}

\markboth{}%
{Shell \MakeLowercase{\textit{et al.}}: A Sample Article Using IEEEtran.cls for IEEE Journals}

\IEEEpubid{}

\IEEEtitleabstractindextext{
\begin{abstract}
With unprecedented rapid development, deep neural networks (DNNs) have deeply influenced almost all fields.
However, their heavy computation costs and model sizes are usually unacceptable in real-world deployment.
Model quantization, an effective weight-lighting technique, has become an indispensable procedure in the whole deployment pipeline.
The essence of quantization acceleration is the conversion from continuous floating-point numbers to discrete integer ones, which significantly speeds up the memory I/O and calculation, \ie addition and multiplication.
However, performance degradation also comes with the conversion because of the loss of precision.
Therefore, it has become increasingly popular and critical to investigate how to perform the conversion and how to compensate for the information loss.
This article surveys the recent five-year progress towards low-bit quantization on DNNs. 
We discuss and compare the state-of-the-art quantization methods and classify them into 8 main categories and 24 sub-categories according to their core techniques.
Furthermore, we shed light on the potential research opportunities in the field of model quantization.
A curated list of model quantization is provided at \hyperlink{https://github.com/Kai-Liu001/Awesome-Model-Quantization}{https://github.com/Kai-Liu001/Awesome-Model-Quantization}.

\end{abstract}

}

\maketitle

\IEEEdisplaynontitleabstractindextext
\IEEEpeerreviewmaketitle

\setlength{\abovedisplayskip}{2pt}
\setlength{\belowdisplayskip}{2pt}

\section{Introduction}

\IEEEPARstart{D}{eep} neural networks (DNNs) are models that extract complex features from data through the interconnection of multiple layers of neurons, demonstrating remarkable performance in various domains, including image recognition~\cite{he2015deepresiduallearningimage, dosovitskiy2021imageworth16x16words, chen2021crossvitcrossattentionmultiscalevision}, processing~\cite{roger2020deepneuralnetworksautomatic, cheng2025hopheterogeneoustopologybasedmultimodal}, and natural language understanding~\cite{liu2019multi, huo2025enhancingnonenglishcapabilitiesenglishcentric, lee2025timecaplearningcontextualizeaugment}.
However, the high performance of DNNs often comes with high computation cost and huge memory space, which limits their application in resource-constrained environments~\cite{krishnamoorthi2018quantizingdeepconvolutionalnetworks}.
Therefore, model compression has grown to be a popular and vital research orientation because of its ability to reduce the parameter size, computation cost, and memory consumption.
Nonetheless, reducing model complexity carries the inherent risk of performance degradation.
Thus, in the field of model compression, achieving a delicate balance between reducing model size and maintaining model accuracy becomes a pivotal challenge.
To address this challenge, researchers have explored various strategies, including but not limited to model pruning, knowledge distillation, innovative model architecture design, and model quantization.     

Quantization is a compression technique in neural network model optimization.
Its key idea is mapping the floating-point weights and/or activation values in the model to low-precision representations, such as integers.
The mapping process, \ie the quantization and de-quantization, could partly slow the inference, while acceleration brought by the low-precision representation is much more obvious.

In model compression, acceleration often comes with performance degradation.
Nonetheless, model quantization achieves the most significant acceleration with minimum degradation.
The Pareto frontier of model quantization is attained with the outstanding talent and hard work of all researchers in this community.
Therefore, we write this quantization survey, aiming to provide a comprehensive understanding and offer insight for future research.

This survey is organized as follows. 
Section~\ref{sec:basic-concepts} introduces the fundamental concepts of model quantization, encompassing the formal definition, elementary quantization approaches, and a foundational taxonomy. 
Subsequently, we categorize recent developments in quantization over the past five years into eight principal groups and twenty-four subcategories based on their core methodologies, as illustrated in Fig.~\ref{fig:taxonomy}. 
A comprehensive analysis and discussion of these techniques are presented in Section~\ref{sec:advanced-topics}, with each subsection corresponding to a specific category of quantization methods. 
Section~\ref{sec:future} outlines several promising directions for future research. 
Due to space constraints, our discussion focuses on the most representative and influential studies in the field. 
Furthermore, we omit extreme quantization techniques (e.g., 1-bit or 1.58-bit quantization) as they involve substantially different methodologies.

\section{Basic Concepts}\label{sec:basic-concepts}

In this section, we present the foundational concepts underlying quantization. 
We begin by formulating the quantization problem, followed by an overview of the principal quantization methodologies. 
Finally, we conclude with classical quantization strategies, highlighting their key characteristics.

\subsection{Formalization of Quantization}
Model quantization involves converting input (weight $W$ or activation $X$) from floating point numbers to a lower precision range through a quantization operator $Q(\cdot)$ which is associated with three parameters, \ie bit-width $b$, scale factor $s$, and zero-point $z$.
The quantization process can be delineated through the following steps:
\begin{equation}\label{eq:quant}
X_{\text{int}} = \text{Clamp}(\text{Round}(\frac{X_{\text{FP}}}{s}) + z, n, p),
\end{equation}
where $X_{\text{int}}$ is the quantized representation, $\text{Clamp}(X, n, p)$ clamps $X$ within $[n,p]$~\cite{nagel2021white}.
The integer grid range $(n, p)$ is determined by the bit-width $b$ , typically covering $\left[-2^{b-1} + 1, 2^{b-1} - 1\right]$ for signed integers and$\left[0, 2^b - 1\right]$ for unsigned integers~\cite{nagel2021white}.
A de-quantization step can be written as:
\begin{equation}
X_{\text{FP}} \approx \hat{X}= s(X_{\text{int}}-z).
\end{equation}

The objective of model quantization is to minimize the evaluation loss, which can be written as
\begin{equation}
\begin{aligned}
\min_{\hat{M}} \quad & \frac{1}{|D_{\text{test}}|} \sum_{(x,y) \in D_{\text{test}}} \mathcal{L}(\hat{M}, x, y), \\
\text{s.t.} \quad & \text{Storage}(\hat{M}) \leq C,
\end{aligned}
\end{equation}
where $M$ and $\hat{M}$ are the FP and quantized model, respectively, $\mathcal{L}$ is the loss function, $ D_{\text{test}} $ is the test set, and $\text{Storage}(\hat{M})$ is the memory overhead.
In some situations, people concentrate on other metrics such as latency, energy, and throughput.
A more commonly used form is to minimize the MSE loss of each layer locally, which can be written as
\begin{equation}
    \min_{Q} ||Q(X)Q(W)-XW||_{2}^{2}.
\end{equation}
This approximation is based on the assumption that lower local quantization loss brings higher model performance.

In summary, classic model quantization aims to seek the best quantization parameters, \ie $b$, $s$, and $z$.
Here, we introduce the three most representative methods.

\subsection{Simple Quantization Methods}
Through Eq.~\eqref{eq:quant}, we can define the bounded range with $(q_{\text{min}}, q_{\text{max}})$, where $q_{\text{min}} = s(n-z)$ and $q_{\text{max}} = s(p-z)$.
Inputs $X$exceeding these bounds are subject to clipping, introducing a clipping error. 
To reduce this, the scale factor can be increased, thus broadening the quantization range.
However, this strategy amplifies the rounding error, which lies in the interval $[-\frac{1}{2}s , \frac{1}{2}s]$.
The methods described below involve different trade-offs between the two types of errors.

\textbf{a) Min-max:}
One way to eliminate clipping errors is to cover the full range of $V$:
\begin{equation}
    q_{\text{min}} = \min V, \quad q_{\text{max}} = \max V.
\end{equation}
However, the Min-max approach is sensitive to extreme values, which are commonly present in many layers and can lead to significant rounding errors.

\textbf{b) Mean Squared Error (MSE):}
To mitigate the influence of outliers and reduce overall quantization error, MSE-based range setting is often employed. 
The quantization limits $(q_{\text{min}} , q_{\text{max}})$ are determined using the subsequent formula:
\begin{equation}
(q_{\text{min}}^{*} , q_{\text{max}}^{*}) = \underset{q_{\text{min}}, q_{\text{max}}}{\text{arg min}} \left\lVert V - Q_{q_{\text{min}}, q_{\text{max}}}(V) \right\rVert_{F}^{2},
\end{equation}
where $\left\lVert \cdot \right\rVert_{F}$ is the Frobenius norm.

\textbf{c) Percentile:}
The quantization range can be determined based on the empirical distribution of the dataset by selecting specific percentiles to define the lower and upper bounds of the range.
For instance, commonly used percentiles include the 99th, 99.9th, or 99.99th percentiles.
When the 99th percentile is employed, the quantization minimum ($q_\text{min}$) is set to the 1st percentile value of the dataset, while the quantization maximum ($q_\text{max}$) corresponds to the 99th percentile value.
This approach effectively mitigates the influence of extreme outliers and enhances quantized model performance and the robustness of the quantization process.

\subsection{Taxonomy of Model Quantization}

\textbf{a) Symmetric and Asymmetric Quantization:} 
Considering the selection of scale factor $s$, which divides real values into a number of partitions:
\begin{equation}
s = \frac{q_{\text{max}} - q_{\text{min}}}{2^b - 1}.
\end{equation}

When $q_{\text{max}}=-q_{\text{min}}$, the quantization range exhibits symmetry around the origin.
Employing symmetric quantization confines the zero-point to 0, thereby reducing the computational cost associated with zero-point offsets~\cite{nagel2021white}. 
However, this symmetry is often not the case, with $q_{\text{max}} \neq -q_{\text{min}}$ indicating an asymmetric clipping range.
Asymmetric quantization, in contrast, allows for a non-zero zero-point. This approach is essential for skewed ranges where symmetric quantization fails to capture the data's variability.
The Min-max quantization scheme, as previously discussed, can serve as an example.

In summary, while symmetric quantization offers computational efficiency, it is not universally applicable. 
Asymmetric quantization provides the necessary flexibility to accurately represent a broader range of data distributions.

\begin{figure*}
\centering
	\includegraphics[width=0.8\linewidth]{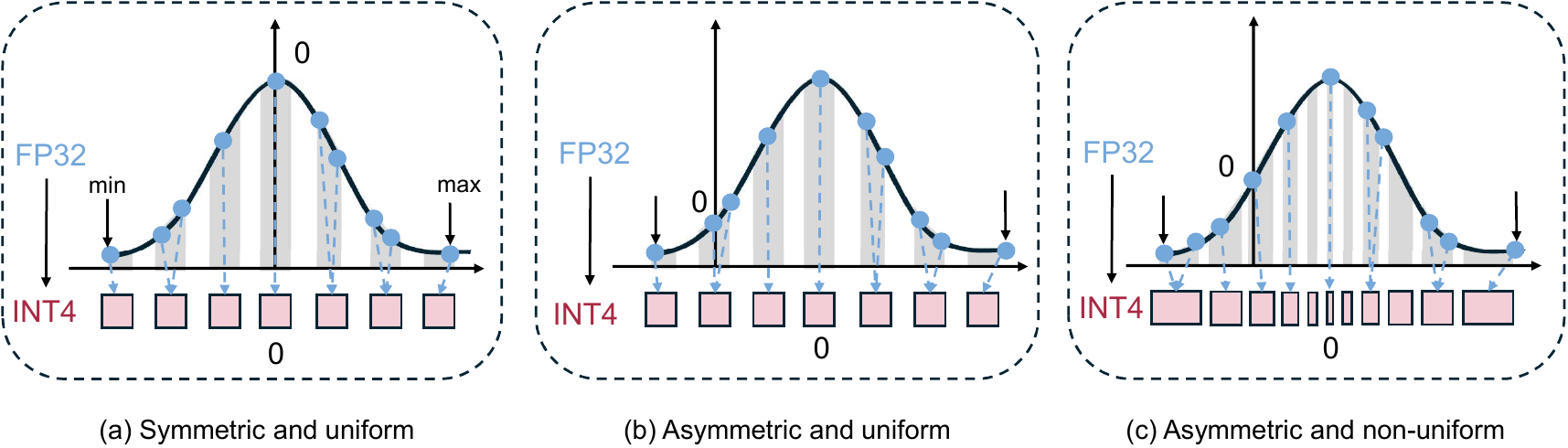}
\vspace{-2mm}
\caption{Illustration of different quantization schemes, including symmetric and asymmetric quantization, uniform and non-uniform quantization.}
\vspace{-2mm}
\end{figure*}

\textbf{b) Uniform and Non-Uniform Quantization:}
The linear quantization operator defined in Eq.~\eqref{eq:quant} is also known as uniform quantization~\cite{nagel2021white}.
Values within the same quantization step size are mapped to the same quantization level according to the rounding function $\text{Int}(\cdot)$.
Considering that more values are distributed around 0, non-uniform quantization is proposed, whose quantization steps are non-uniformly spaced. 
It can be defined as:
\begin{equation}
q(x) = X_i, \text{ if } x \in [\Delta_i, \Delta_{i+1}).
\end{equation}
Specifically, for a real number $x$ that lies between quantization steps $i$ and $ i+1 $, the quantizer $q$ maps it to the corresponding level $X_i$. Neither the $X_i$ levels nor the indices $i$ are uniformly spaced.
A typical non-uniform quantization uses a logarithmic distribution, where the quantization steps and levels increase exponentially.
 
In recent years, non-uniform quantization has made breakthrough achievements, especially in the field of extreme quantization. However, it is difficult to achieve the acceleration effect brought by low bit-width numerical matrix operations in uniform linear quantization. Therefore, on hardware with limited resources, models based on non-uniform quantization are difficult to bring practical benefits.

In summary, non-uniform quantization enhances precision by targeting critical value regions according to data distribution. Yet, its deployment on standard computing hardware, including GPUs and CPUs, poses challenges. Consequently, uniform quantization remains the most common and efficient approach for quantization.

\textbf{c) Static And Dynamic Quantization:}
The time when to decide the clipping range  $(q_{\text{min}}, q_{\text{max}})$ is crucial and can be divided into static and dynamic quantization~\cite{nagel2021white}.

In dynamic quantization, scale factor and zero point are calculated in real time, allowing the model to adapt precisely to variable input data and potentially enhancing accuracy.
However, this adaptive process incurs significant computational overhead due to the continuous estimation of the range during inference.
Instead, the clipping range is pre-calculated in static quantization, which introduces no computational overhead with reduced accuracy.
In summary, dynamic quantization enhances accuracy at a computational cost, while static quantization is efficient but less accurate.

\textbf{d) Quantization Granularity:}
Quantization granularity refers to the level of quantization parameters applied in quantization, impacting accuracy, computation, and storage. 
Coarser granularities simplify implementation but may compromise accuracy due to a less precise range of data across layer filters.
The following four granularities are arranged from coarse to fine.

\textbf{Per-tensor:} Also known as per-layer, this method uses a set of quantization parameters for each tensor, with the same clipping range applied to all convolution filters. It's the most common choice due to simplified hardware implementation.
\textbf{Per-group:} This method divides the model into groups, typically in CNNs, assigning each group a unique quantization parameter. It offers a balance between flexibility and computational efficiency compared to per-tensor and per-channel methods.
\textbf{Per-channel:} Here, unique parameters are assigned to each channel, improving accuracy when weight distributions vary, and removing the need for global rescaling.
\textbf{Per-token:} Applied to language models, this approach assigns quantization parameters to individual tokens (words or characters), offering fine-grained precision.

\begin{figure}
\centering
\includegraphics[width=\linewidth]{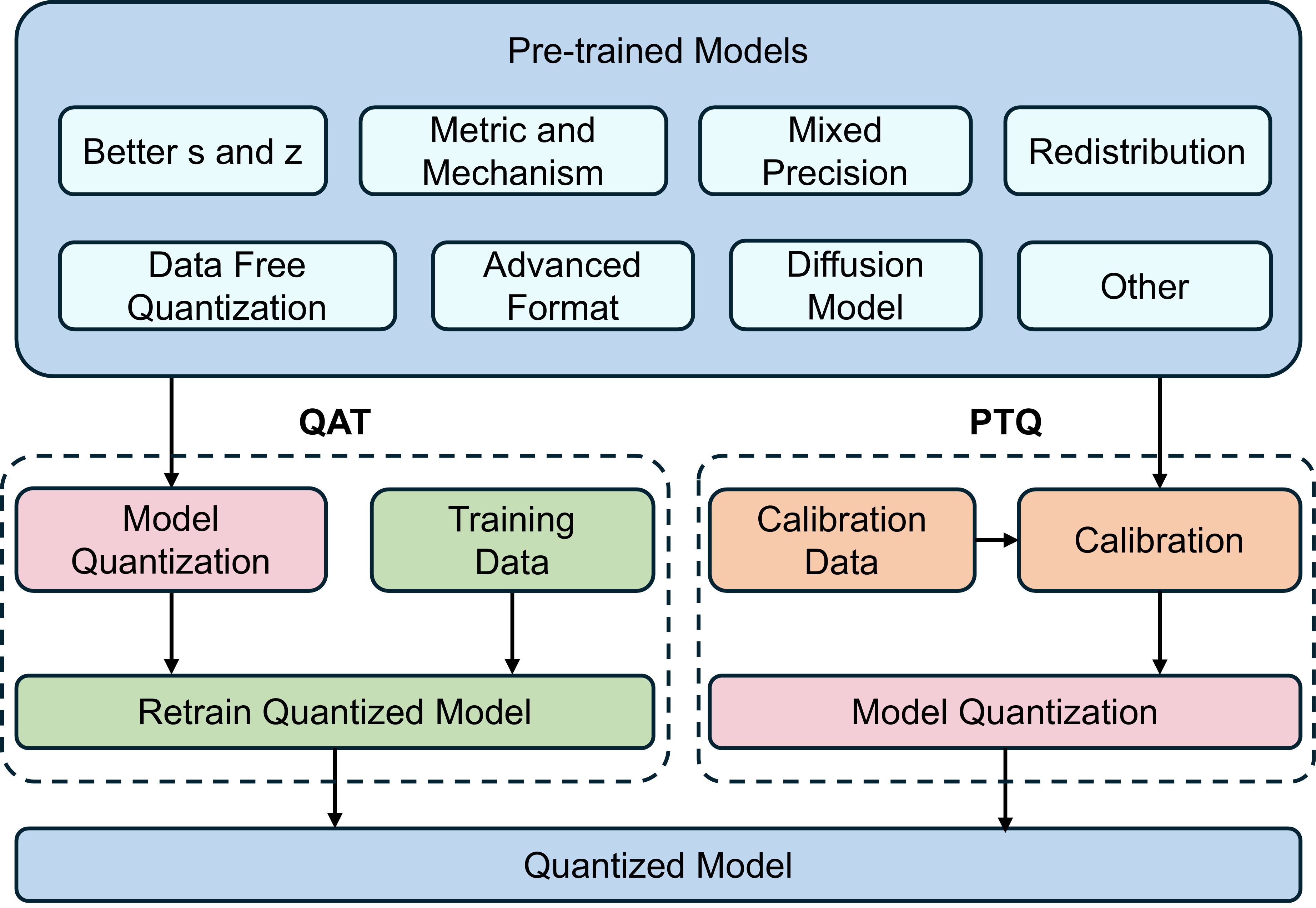}
\vspace{-2mm}
\caption{QAT usually takes large datasets to retrain the weights and quantizers' parameters, while PTQ mostly only leverages the calibration data to obtain the quantizer parameters.}
\label{fig:ptq-qat}
\vspace{-2mm}
\end{figure}

\textbf {e) PTQ and QAT:}
In the early stage of model quantization, quantization is only employed on the pre-trained models.
Researchers leverage efficient algorithms to determine the quantizer parameters of each layer with a small calibration set, which is called post-training quantization (PTQ).
With the proposal of STE~\cite{bengio2013estimating}, quantization-aware training (QAT) attracts people's attention and significantly improves the performance.
STE allows gradients back propagation and therefore increases the global performance.
QAT usually takes training on a larger training set with adjustments to both model parameters and quantizer parameters.
In recent research, more methods employ STE in PTQ to optimize the quantizer parameters.
Efficient algorithms are also proposed to adjust the model parameters.
The distinction between PTQ and QAT is becoming smaller.
In the following section, we present the inner relationship between these two quantization methods and recent advanced methods, as shown in Fig.~\ref{fig:ptq-qat}.

\begin{figure*}
\centering
\includegraphics[width=\linewidth]{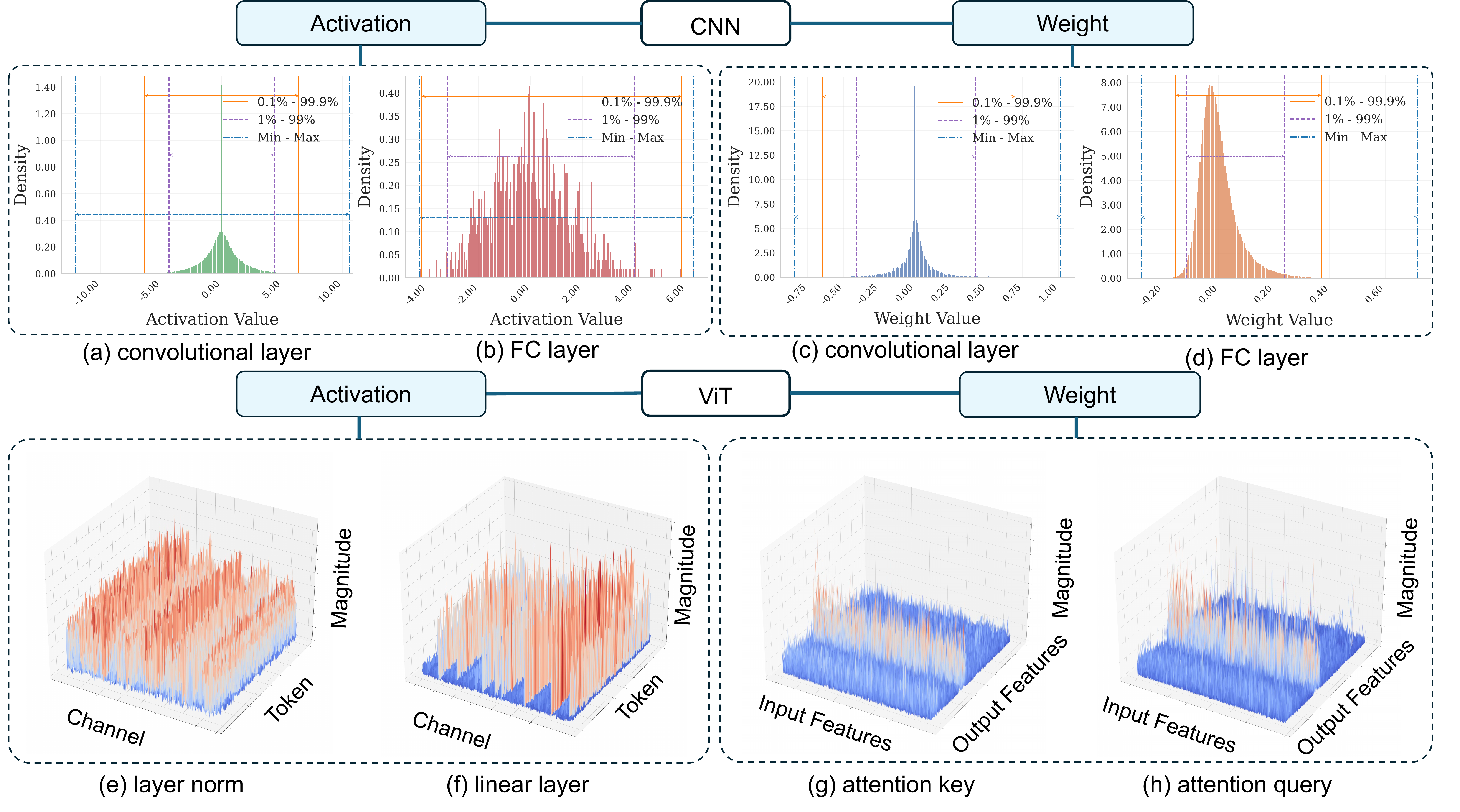}
\vspace{-8mm}
\caption{Distribution of activation and weight in common models. For CNN, the distributions of activations and weights are mostly symmetrical and similar to normal distribution. For ViT, the situation is different. Both activation and weights show obvious correlations along the channel.}
\vspace{-4mm}
\label{fig:cnn-vit}
\end{figure*}

\forestset{
  main'/.style={
    l sep=0mm,
    anchor=west,
  },
  root'/.style={root,
    anchor=west,
    edge path={
     \noexpand\path[\forestoption{edge}]
     ($(!u.east)!.0!(!.west)$) ++(0.65, 0) |- (!.west);
    },
  },
  parent'/.style={parent, 
    anchor=west,
    calign=child edge,
    l sep=0.2cm
  },
}

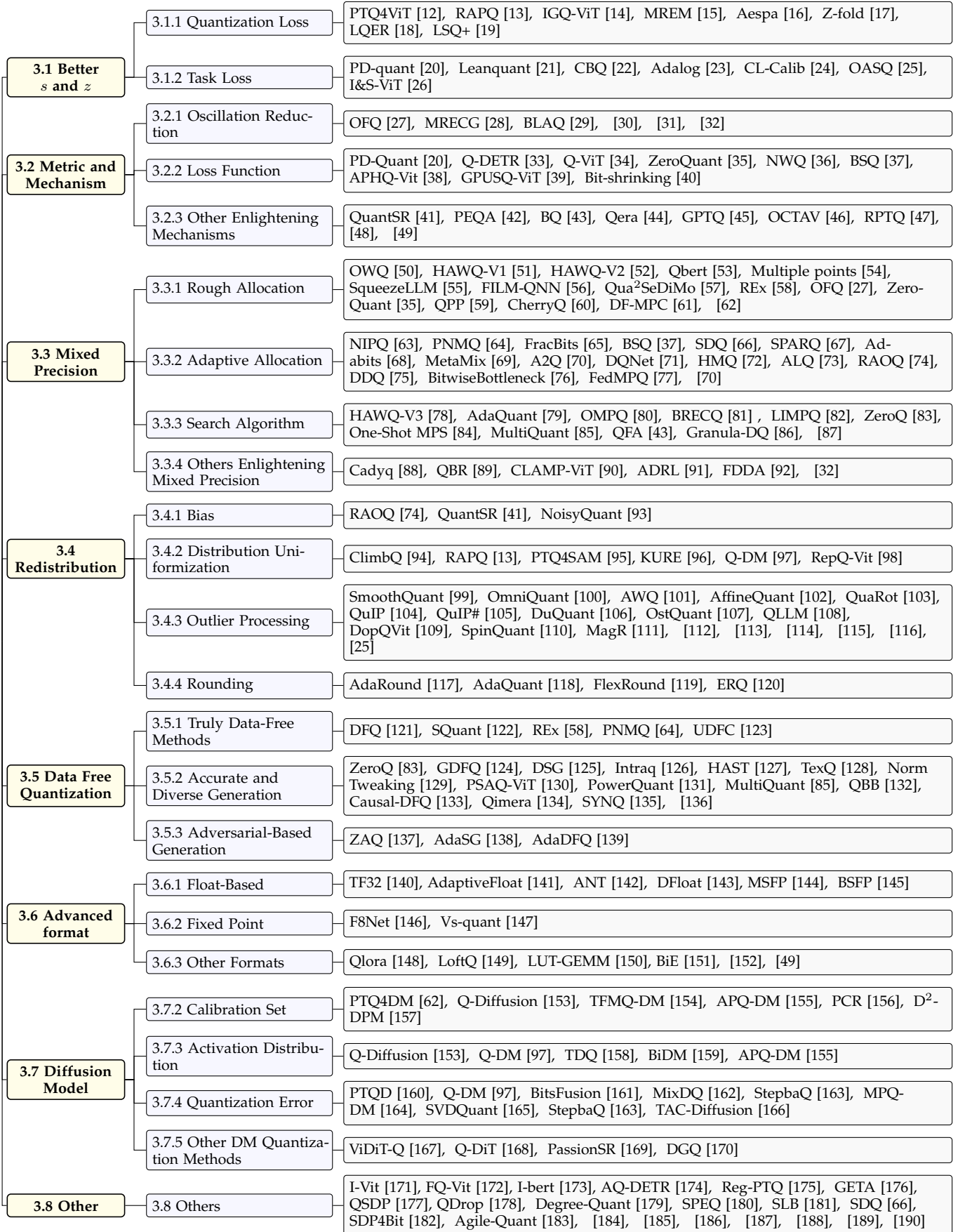
\begin{figure*}[ht!]
    \begin{adjustwidth}{0cm}{0cm}
    \hspace{-0.85cm} 
    \centering
\begin{forest}
    for tree={
        forked edges,
        text centered,
        grow'=east,
        reversed=true,
        font=\footnotesize,
        rectangle, /tikz/align=left, anchor=base west, tier/.pgfmath=level(),
        rounded corners,
    }
    [, main'
        [\ref{Better zero-point and scale} Better $s$ and $z$, root', calign=child, calign child=2 
            [\ref{sec:quant-loss} Quantization Loss, parent'
                [
                    PTQ4ViT~\cite{yuan2024ptq4vit}\text{, }  
                    RAPQ~\cite{yao2022rapqrescuingaccuracypoweroftwo}\text{, }
                    IGQ-ViT~\cite{moon2024instance}\text{, }
                    MREM~\cite{bai2021efficientposttrainingquantizationpretrained}\text{, }
                    Aespa~\cite{kim2024towards}\text{, }
                    Z-fold~\cite{jeon2023a}\text{, }
                    LQER~\cite{zhang2024lqerlowrankquantizationerror}\text{, }
                    LSQ+~\cite{bhalgat2020lsqimprovinglowbitquantization}
                    , child
                ]
            ]
            [\ref{sec:task-loss} Task Loss, parent'
                [
                    PD-quant~\cite{liu2022pd}\text{, }
                    Leanquant~\cite{zhang2024leanquant}\text{, }
                    CBQ~\cite{ding2025cbq}\text{, }
                    Adalog~\cite{wu2024adalog}\text{, }
                    CL-Calib~\cite{shang2024clcalib}\text{, }
                    OASQ~\cite{ma2024outlieraware}\text{, }
                    I\&S-ViT~\cite{zhong2024isvitinclusivestable}
                    , child
                ]
            ]
        ]
        [\ref{sec:metric-and-mechanism} Metric and Mechanism, root'
            [\ref{sec:oscillation-reduction} Oscillation Reduction, parent'
                [
                    OFQ~\cite{liu2023oscillation}\text{, }
                    MRECG~\cite{ma2023solving}\text{, }
                    BLAQ~\cite{ma2024one}\text{, }
                    ~\cite{lee2022toward}\text{, }
                    ~\cite{nagel2022overcoming}\text{, }
                    ~\cite{tang2024retraining}
                    , child
                ]
            ]
            [\ref{Loss Function} Loss Function, parent'
                [
                    PD-Quant~\cite{liu2022pd}\text{, }
                    Q-DETR~\cite{xu2023q}\text{, }
                    Q-ViT~\cite{li2022qvit}\text{, }
                    ZeroQuant~\cite{yao2022zeroquant}\text{, }
                    NWQ~\cite{zheng2022leveraging}\text{, }
                    BSQ~\cite{yang2021bsq}\text{, }
                    APHQ-Vit~\cite{wu2025aphq}\text{, }
                    GPUSQ-ViT~\cite{yu2023boost}\text{, }
                    Bit-shrinking~\cite{10203557}
                    , child
                ]
            ]
            [\ref{Other Enlightening Update Mechanisms} Other Enlightening Mechanisms, parent'
                [       
                    QuantSR~\cite{qin2023quantsr}\text{, }
                    PEQA~\cite{kim2023memory}\text{, }
                    BQ~\cite{bai2021batchquant}\text{, }
                    Qera~\cite{zhang2025qera}\text{, }
                    GPTQ~\cite{frantar2023gptq}\text{, }
                    OCTAV~\cite{sakr2022optimalclippingmagnitudeawaredifferentiation}\text{, }
                    RPTQ~\cite{yuan2023rptq}\text{, }
                    ~\cite{fu2024quantizationtears}\text{, }
                    ~\cite{wang2022learnable}
                    , child 
                    ]
            ]
        ]
        [\ref{Mixed-Precision} Mixed Precision, root', calign=child, calign child=2
            [\ref{Rough Bit-width Allocation} Rough Allocation, parent'
                [
                    OWQ~\cite{lee2023owq}\text{, }
                    HAWQ-V1~\cite{dong2019hawq}\text{, }
                    HAWQ-V2~\cite{dong2020hawq}\text{, }
                    Qbert~\cite{shen2019qbert}\text{, }
                    Multiple points~\cite{liu2021post}\text{, }
                    SqueezeLLM~\cite{kim2024squeezellm}\text{, }
                    FILM-QNN~\cite{sun2022film}\text{, }
                    Qua$^2$SeDiMo~\cite{mills2024qua}\text{, }
                    REx~\cite{yvinec2023rex}\text{, }
                    OFQ~\cite{liu2023oscillation}\text{, }
                    ZeroQuant~\cite{yao2022zeroquant}\text{, }
                    QPP~\cite{kryzhanovskiy2021qpp}\text{, }
                    CherryQ~\cite{cui2024cherry}\text{, }
                    DF-MPC~\cite{chen2023data}\text{, }
                    ~\cite{shang2023ptqdm}
                    , child
                ]
            ]
            [\ref{Adaptive Bit-width Allocation} Adaptive Allocation, parent'
                [
                    NIPQ~\cite{shin2023nipqnoiseproxybasedintegrated}\text{, }
                    PNMQ~\cite{chikin2022data}\text{, }
                    FracBits~\cite{yang2020fracbits}\text{, } 
                    BSQ~\cite{yang2021bsq}\text{, } 
                    SDQ~\cite{huang2022sdq}\text{, }
                    SPARQ~\cite{shomron2021post}\text{, }
                    Adabits~\cite{jin2020adabits}\text{, }
                    MetaMix~\cite{kim2024metamix}\text{, }
                    A2Q~\cite{zhu2023rm}\text{, }
                    DQNet~\cite{liu2022instance}\text{, }
                    HMQ~\cite{habi2020hmqhardwarefriendlymixed}\text{, }
                    ALQ~\cite{qu2020adaptive}\text{, }
                    RAOQ~\cite{zhang2024reshape}\text{, }
                    DDQ~\cite{zhang2021differentiable}\text{, }
                    BitwiseBottleneck~\cite{zhou2021optimizing}\text{, }
                    FedMPQ~\cite{chen2024mixed}\text{, }
                    ~\cite{zhu2023rm}
                    , child
                ]
            ]
            [\ref{Optimized Mixed Precision Search Algorithm} Search Algorithm, parent'
                [   
                    HAWQ-V3~\cite{yao2021hawqv3}\text{, }
                    AdaQuant~\cite{hubara2021accurate}\text{, }
                    OMPQ~\cite{ma2023ompq}\text{, }
                    BRECQ~\cite{li2021brecq} \text{, }
                    LIMPQ~\cite{tang2022mixed}\text{, }
                    ZeroQ~\cite{cai2020zeroq}\text{, }
                    One-Shot MPS~\cite{koryakovskiy2023one}\text{, }
                    MultiQuant~\cite{xu2022multiquant}\text{, }
                    QFA~\cite{bai2021batchquant}\text{, }
                    Granula-DQ~\cite{wang2024thinkinggranularitydynamicquantization}\text{, }
                    ~\cite{liu2021posttrainingquantizationvisiontransformer}
                    , child
                ]
            ]
            [\ref{other} Others Enlightening Mixed Precision, parent'
                [   
                    Cadyq~\cite{hong2022cadyqcontentawaredynamicquantization}\text{, }
                    QBR~\cite{sun2022entropy}\text{, }
                    CLAMP-ViT~\cite{ramachandran2024clamp}\text{, }
                    ADRL~\cite{ning2021simple}\text{, }
                    FDDA~\cite{zhong2022finegraineddatadistributionalignment}\text{, }
                    ~\cite{tang2024retraining}
                    , child
                ]
            ]
        ]
        [\ref{Redistribution} Redistribution, root', calign=child, calign child=2
            [\ref{Bias} Bias, parent'
                [
                    RAOQ~\cite{zhang2024reshape}\text{, }
                    QuantSR~\cite{qin2023quantsr}\text{, }
                    NoisyQuant~\cite{liu2023noisyquantnoisybiasenhancedposttraining} 
                    , child
                ]
            ]
            [\ref{Distribution uniformization} Distribution Uniformization, parent'
                [
                    ClimbQ~\cite{chen2022climbq}\text{, }
                    RAPQ~\cite{yao2022rapqrescuingaccuracypoweroftwo}\text{, }
                    PTQ4SAM\cite{lv2024ptq4sam}\text{,}
                    KURE~\cite{shkolnik2020robustquantizationmodelrule}\text{, }
                    Q-DM~\cite{li2023qdm}\text{, }
                    RepQ-Vit~\cite{li2023repq}
                    , child
                ]
            ]
            [\ref{Outlier Processing} Outlier Processing, parent'
                [
                    SmoothQuant~\cite{xiao2024smoothquantaccurateefficientposttraining}\text{, }
                    OmniQuant~\cite{shao2024omniquantomnidirectionallycalibratedquantization}\text{, }
                    AWQ~\cite{lin2023awq}\text{, }
                    AffineQuant~\cite{ma2024affinequant}\text{, }
                    QuaRot~\cite{ashkboos2024quarotoutlierfree4bitinference}\text{, }
                    QuIP~\cite{chee2023quip}\text{, }
                    QuIP\#~\cite{tseng2024quip}\text{, }
                    DuQuant~\cite{lin2024duquantdistributingoutliersdual}\text{, }
                    OstQuant~\cite{hu2025ostquantrefininglargelanguage}\text{, }
                    QLLM~\cite{liu2024qllmaccurateefficientlowbitwidth}\text{, }
                    DopQVit~\cite{yang2024dopqvitdistributionfriendlyoutlierawareposttraining}\text{, }
                    SpinQuant~\cite{liu2025spinquantllmquantizationlearned}\text{, }
                    MagR~\cite{zhang2024magrweightmagnitudereduction}\text{, }
                    ~\cite{lin2024qserve}\text{, }
                    ~\cite{xi2023trainingtransformers4bitintegers}\text{, }
                    ~\cite{wei2023outliersuppressionaccuratequantization}\text{, }
                    ~\cite{martinez2021permutequantizefinetuneefficient}\text{, }
                    ~\cite{Nie2022RedistributionOW}\text{, }
                    ~\cite{ma2024outlieraware}
                    , child
                ]
            ]
            [\ref{Rounding} Rounding, parent'
                [
                    AdaRound~\cite{nagel2020downadaptiveroundingposttraining}\text{, }
                    AdaQuant~\cite{hubara2020improvingposttrainingneural}\text{, }
                    FlexRound~\cite{lee2024flexroundlearnableroundingbased}\text{, }
                    ERQ~\cite{zhong2024erq}
                    , child
                ]
            ]
        ]
        [\ref{sec:data-free-quantization} Data Free Quantization, root', calign=child, calign child=2
            [\ref{sec:truly-data-free-methods} Truly Data-Free Methods , parent'
                [
                    DFQ~\cite{nagel2019data}\text{, }
                    SQuant~\cite{guo2022squant}\text{, }
                    REx~\cite{yvinec2023rex}\text{, }
                    PNMQ~\cite{chikin2022data}\text{, }
                    UDFC~\cite{bai2023unified}
                    , child
                ]
            ]
            [\ref{sec:accurate-and-diverse-generation} Accurate and Diverse Generation, parent'
                [
                     ZeroQ~\cite{cai2020zeroq}\text{, }
                     GDFQ~\cite{xu2020generativelowbitwidthdatafree}\text{, }
                     DSG~\cite{zhang2021diversifyingsamplegenerationaccurate}\text{, }
                     Intraq~\cite{zhong2022intraqlearningsyntheticimages}\text{, }
                     HAST~\cite{li2023hardsamplematterslot}\text{, }
                     TexQ~\cite{chen2023texq}\text{, }
                     Norm Tweaking~\cite{li2023normtweakinghighperformancelowbit}\text{, }
                     PSAQ-ViT~\cite{li2022psaqvit}\text{, }
                     PowerQuant~\cite{yvinec2023powerquantautomorphismsearchnonuniform}\text{, }
                     MultiQuant~\cite{xu2022multiquant}\text{, }
                     QBB~\cite{bulat2024qbb}\text{, }
                     Causal-DFQ~\cite{shang2023causal}\text{, }
                     Qimera~\cite{choi2021qimera}\text{, }
                     SYNQ~\cite{kimsynq}\text{, }
                     ~\cite{fan2024data}
                    , child
                ]
            ]
            [\ref{sec:adversarial-based-generation} Adversarial-Based Generation, parent'
                [
                     ZAQ~\cite{liu2021zeroshotadversarialquantization}\text{, }
                     AdaSG~\cite{Qian_Wang_Hong_Wang_2023}\text{, }
                     AdaDFQ~\cite{qian2023adaptive}
                    , child
                ]
            ]
        ]
        [\ref{Advanced format} Advanced format, root', calign=child, calign child=2
            [\ref{FP-based} Float-Based, parent'
                [
                    TF32~\cite{chmiel2020neuralgradientsnearlognormalimproved}\text{,}
                    AdaptiveFloat~\cite{tambe2020algorithm}\text{, }
                    ANT~\cite{guo2022antexploitingadaptivenumerical}\text{, }
                    DFloat~\cite{zhang202570size100accuracy}\text{,}
                    MSFP~\cite{darvishrouhani2020pushing}\text{, }
                    BSFP~\cite{lo2023block}
                    , child
                ]
            ]
            [\ref{Fixed Point} Fixed Point, parent'
                [   
                    F8Net~\cite{jin2022f8net}\text{, }
                    Vs-quant~\cite{dai2021vs}
                    , child
                ]
            ]
            [\ref{other formats} Other Formats, parent'
                [   
                    Qlora~\cite{dettmers2023qloraefficientfinetuningquantized}\text{, }
                    LoftQ~\cite{li2023loftq}\text{, }
                    LUT-GEMM~\cite{park2024lutgemmquantizedmatrixmultiplication}\text{,}
                    BiE~\cite{zou2024bie}\text{,}
                    ~\cite{dotzel2024learning}\text{,}
                    ~\cite{wang2022learnable}
                    , child
                ]
            ]
        ]
        [\ref{sec:DM} Diffusion Model, root'
            [\ref{sec:Calibration} Calibration Set, parent'
                [
                    PTQ4DM~\cite{shang2023ptqdm}\text{, }
                    Q-Diffusion~\cite{li2023qdiffusion}\text{, }
                    TFMQ-DM~\cite{Huang_2024_CVPR}\text{, }
                    APQ-DM~\cite{wang2024towards}\text{, }
                    PCR~\cite{tang2024post}\text{, }
                    D$^{2}$-DPM~\cite{zeng2025d}
                    , child
                ]
            ]
            [\ref{Challenges Concerning Activation Distribution} Activation Distribution, parent'
                [
                    Q-Diffusion~\cite{li2023qdiffusion}\text{, }
                    Q-DM~\cite{li2023qdm}\text{, }
                    TDQ~\cite{so2023temporal}\text{, }
                    BiDM~\cite{zheng2024bidm}\text{, }
                    APQ-DM~\cite{wang2024towards}
                    , child
                ]
            ]
            [\ref{Quantization error} Quantization Error, parent'
                [
                    PTQD~\cite{he2023ptqd}\text{, }
                    Q-DM~\cite{li2023qdm}\text{, }
                    BitsFusion~\cite{sui2024bitsfusion}\text{, }
                    MixDQ~\cite{zhao2024mixdq}\text{, }
                    StepbaQ~\cite{chen2024stepbaq}\text{, }
                    MPQ-DM~\cite{feng2024mpq}\text{, }
                    SVDQuant~\cite{li2024svdqunat}\text{, }
                    StepbaQ~\cite{chen2024stepbaq}\text{, }
                    TAC-Diffusion~\cite{yao2024timestep}
                    , child
                ]
            ]
            [\ref{sec:other-DM} Other DM Quantization Methods, parent'
                [
                   ViDiT-Q~\cite{zhao2024vidit}\text{, }
                   Q-DiT~\cite{chen2024q}\text{, }
                   PassionSR~\cite{zhu2024passionsr}\text{, }
                   DGQ~\cite{ryu2025dgq}
                    , child
                ]
            ]
        ]
        [\ref{sec:other} Other, root', calign=child, calign child=2
            [\ref{sec:other} Others, parent'
                [
                    I-Vit~\cite{li2023ivitintegeronlyquantizationefficient}\text{,}
                    FQ-Vit~\cite{lin2023fqvitposttrainingquantizationfully}\text{,}
                    I-bert~\cite{kim2021ibertintegeronlybertquantization}\text{,}
                    AQ-DETR~\cite{wang2024aq}\text{, }
                    Reg-PTQ~\cite{10654966}\text{, }
                    GETA~\cite{qu2025automaticjointstructuredpruning}\text{,}
                    QSDP~\cite{markov2023quantizeddistributedtraininglarge}\text{,}
                    QDrop~\cite{wei2023qdrop}\text{, }
                    Degree-Quant~\cite{tailor2021degree}\text{, }
                    SPEQ~\cite{boo2021stochastic}\text{, }
                    SLB~\cite{yang2020searchinglowbitweightsquantized}\text{, }
                    SDQ~\cite{huang2022sdq}\text{, }
                    SDP4Bit~\cite{jia2024sdp4bit4bitcommunicationquantization}\text{, }
                    Agile-Quant~\cite{shen2025agilequantactivationguidedquantizationfaster}\text{, }
                    ~\cite{tianqi2023towards}\text{, }
                    ~\cite{dettmers20228bitoptimizersblockwisequantization}\text{, }
                    ~\cite{lee2021networkquantizationelementwisegradient}\text{, }
                    ~\cite{chu2023makerepvgggreateragain}\text{, }
                    ~\cite{fan2021training}\text{, }
                    ~\cite{heo2025rethinkingchanneldimensionsisolate}\text{,}
                    ~\cite{wang2022leveraging}
                    , child
                ]
            ]
        ]
    ]
\end{forest}
\vspace{-4mm}
\caption{Taxonomy for recent quantization methods. We classify 179 papers into 8 categories and further into 24 sub-categories.}
\label{fig:taxonomy}
\end{adjustwidth}
\vspace{-6mm}
\end{figure*}

\section{Advanced topics}\label{sec:advanced-topics}
In this section, we surveyed 179 quantization papers and classified them into 8 main categories (Fig.~\ref{fig:taxonomy}), according to their main techniques.
The 8 main categories are 
better $s$ and $z$, 
metric and mechanism, 
mixed precision,
redistribution,
data-free quantization, 
advanced format, 
diffusion model, 
and other.
Some papers are classified into multiple categories.

\subsection{Better $s$ and $z$}\label{Better zero-point and scale}
As previously mentioned, the quantization range and scale factors are fundamental elements in a quantization scheme, and their determination establishes the overarching framework for the entire quantization process. 
Recent works~\cite{bai2021batchquant, yuan2024ptq4vit} have indicated that common quantization metrics, such as Mean Square Error (MSE) and Exponential Moving Average (EMA), are not accurate in determining optimal scale factors and quantization ranges. 
Therefore, in recent literature, a multitude of innovative approaches have been proposed regarding how to determine the quantization range and scale factors. 
Next, we summarize the main approaches from two perspectives.

\subsubsection{Quantization Loss}\label{sec:quant-loss}

PTQ4ViT~\cite{yuan2024ptq4vit} introduced a dual uniform quantization strategy, utilizing two separate quantization ranges, $R_{1}$ and $R_{2}$, each with its own scaling factor $\Delta R_1$ and $\Delta R_2$. 
This approach allows for the use of different scaling factors for positive and negative values, as well as for extremely high and low values.
RAPQ~\cite{yao2022rapqrescuingaccuracypoweroftwo} determines the scaling factor groups through an iterative approach, with these factors typically being powers of 2 to facilitate hardware implementation.
RAPQ solves for the optimal solutions of two regularization terms to determine the scaling factor groups for weights and activations. 
This method dynamically adjusts the Power-of-Two quantization scale for the entire network, rather than statically determining for each layer.

For each input instance, IGQ-ViT~\cite{moon2024instance} dynamically divides the channels of the activation map into multiple groups, ensuring that the activation values within each group have similar statistical characteristics. 
Then, by minimizing the distance between the min and max values of each group's activations and the upper and lower bounds of the quantizer, the scaling parameters $s$ and zero point $z$ for each group are optimized to achieve a minimal quantization error.

In addition to existing out-channel wise scaling factors $\alpha \in \mathbb{R}^{n \times 1}$, Z-fold~\cite{jeon2023a} introduced new scaling factor $\zeta \in \mathbb{R}^{m \times 1}$ corresponding to the in-channel.
To achieve a bi-directional step size matrix $S=\zeta \cdot \alpha^{\top}$, Z-FOLD utilizes alternating least squares to iteratively update $\alpha$ and  $\zeta$ in order to minimize the quantization error.

MREM~\cite{bai2021efficientposttrainingquantizationpretrained} divides the Pre-trained Language Model into multiple modules, with each module containing several Transformer layers, and jointly optimizing all layers within the module to better capture the inter-layer correlations. 
The optimization objective is:
\begin{align}
\min_{w_n, s_n} \ell^{(n)} = \sum_{l \in [l_n, l_{n+1})} \| \hat{f}_l - f_l \|_2^2,
\end{align}
where $w_n$ and $s_n$ are the learnable parameters and quantization step sizes in the nth module, $ \hat{f}_l $ is the quantized FFN output, and $f_l$ is the full-precision output.
Aespa~\cite{kim2024towards} also used this to optimize $S$.
LQER~\cite{zhang2024lqerlowrankquantizationerror} reconstructs the quantization error matrix to restore the performance of the model. 
Specifically, it decomposes the weight matrix into a low-precision, high-rank quantized matrix and a high-precision, low-rank error approximation matrix.
And LSQ+~\cite{bhalgat2020lsqimprovinglowbitquantization} introduces a general asymmetric quantization scheme with trainable $s$ and $z$.

\subsubsection{Task Loss}\label{sec:task-loss}
PD-Quant~\cite{liu2022pd} pointed out that methods using local metrics to search for quantization scaling factors consider only the local changes in the activation of each layer.
That may cause the scaling factors not to align with those optimized by task loss. 
To address this, PD-Quant considers global information by comparing the prediction differences between the full-precision and the quantized models, and optimizes the scaling factors following the formula:
\begin{equation}
    \arg \min _{S_{a}} \mathcal{L}_{P D}\left(f_{l}\left(\tilde{A}_{l-1}\right), f_{l+1}\left(L_{l}^{q}\left(\tilde{A}_{l-1}\right)\right)\right),
\end{equation}
where $S_a$ denotes scale for activation of layer $L_{l}^{q}$, and $f_l(\cdot)$ refers to a part of the FP network mapping input $\tilde{A}_{l-1}$ to FP output~\cite{liu2022pd}.
$f_{l+1}(\cdot)$ is the part of the FP network mapping the output of $L_{l}^{q}$ to the final output~\cite{liu2022pd}.

CL-Calib~\cite{shang2024clcalib} designs a contrastive learning loss function, which aims to reduce the distance between positive sample pairs while increasing the distance between negative sample pairs.
Positive sample pairs and negative sample pairs are generated by the FP activations and quantized activations from the same and different input samples, respectively.
By minimizing the contrastive learning loss function, adjustments to $s$ and $z$ can be realized, enhancing the mutual information between quantized activations and FP activations~\cite{shang2024clcalib}.

LeanQuant~\cite{zhang2024leanquant} uses a loss-error-aware quantization grid. For affine quantization, LeanQuant optimizes $s$ and $z$ to minimize the total loss error during the quantization process. Specifically, LeanQuant performs a discrete search in a restricted space to find the optimal $s$ and $z$, thus maintaining a uniform distribution of the quantization grid while minimizing the impact of quantization errors. 
CBQ~\cite{ding2025cbq} is designed to optimize layers that have significant dependencies. By employing a sliding window approach, CBQ simultaneously optimizes multiple blocks within the window, and there is overlap between adjacent sliding windows to ensure connectivity between blocks. 
The optimization goal is to reduce the task error between the FP and quantized models.
Adalog~\cite{wu2024adalog} searches for hyperparameters by progressively narrowing down the search space. 
It can more accurately locate the optimal hyperparameters while maintaining linear complexity.

\subsection{Metric and Mechanism}\label{sec:metric-and-mechanism}
In this section, we focus on the metrics and mechanisms used in the QAT process.
We divide the related papers into three categories.
The first category is about the loss functions, which leverage novel and effective metrics to achieve better fine-tuning.
The second category is about oscillation, which represents misconvergence of weights on quantization boundaries.
The rest papers provide a different perspective and they are classified into the last category.

\subsubsection{Oscillation Reduction}\label{sec:oscillation-reduction}

Oscillation exists widely in the QAT process, which refers to the random fluctuation of weights between two quantization levels rather than converging to a stable value. 
This phenomenon originates from the STE near quantization boundaries, causing the gradient to change direction abruptly when crossing quantization thresholds, which leads to weights oscillating between adjacent quantization levels.

When training a full-precision neural network, we expect the weights to change slowly as they approach convergence.
So that the output of each layer is expected to be stable across iterations, making the mean and variance of BN layers good estimates. 
However, oscillations lead to rapid changes in integer weights, causing significant distribution shifts between iterations.
These sudden and substantial changes in output distribution can result in a significant decrease in accuracy.
The following researches are dedicated to solving the oscillation problem~\cite{lee2022toward,nagel2022overcoming,liu2023oscillation,tang2024retraining,ma2023solving,ma2024one}.

Hysteresis quantization is proposed, where the quantization value increases only when the increase is large enough to surpass the current quantized value, vice versa, reducing frequent switching between two quantization values~\cite{lee2022toward} .
A regularization term is introduced that encourages latent weights to stay closer to the center of the quantization bins rather than the edges~\cite{nagel2022overcoming}. 
It also tracks the oscillation frequency of each weight during training.
If any weight's oscillation frequency exceeds a threshold, that weight is frozen until the end of training.
It is notable that, opposite to~\cite{nagel2022overcoming}, OFQ~\cite{liu2023oscillation} chooses to freeze weights that are far from quantization thresholds and only concerns those close to the thresholds.
Targeting ViT, OFQ also reorders the multiplication operation of queries and keys in the self-attention module, decoupling the oscillations.

In the weight-sharing quantization model, weight oscillations originate from the bit-width interference problem, which causes the same weights to be quantized to different quantization levels.
~\cite{tang2024retraining} proposes a dynamic bit-width schedule to freeze the most unstable bit-widths, thereby reducing weight oscillation during quantization.

Methods mentioned above are all set in the background of QAT, however, MRECG~\cite{ma2023solving} reveals that the same oscillation issues exist in PTQ.
It smooths oscillations by jointly optimizing adjacent modules with significant capacity differences.

Additionally, there is another oscillation issue, which refers to the sawtooth changes in gradient direction during the gradient descent learning process.
BLAQ~\cite{ma2024one} tackles this issue by utilizing a one-step forward search to determine the next trial gradient, followed by a backtrack update to revert to the current step. 
Both the FP and quantized parameters are updated with the current and trial gradients.

\subsubsection{Loss Function}\label{Loss Function}
Recent works~\cite{bai2021batchquant, yuan2024ptq4vit} have indicated that traditional loss functions, such as MSE and Exponential Moving Average (EMA), such as MSE and Cross-Entropy (CE), may not be sufficient to guide model learning.Researchers~\cite{liu2022pd,xu2023q,li2022qvit,yao2022zeroquant,zheng2022leveraging,yang2021bsq,yu2023boost,frumkin2023jumping} are committed to adjusting loss functions to meet different demands of quantization.

For example, PD-Quant~\cite{liu2022pd} introduces predictive difference loss.
The loss is calculated by comparing the output predictions of the full-precision model and the quantized model, rather than just the differences of each layer's activation before and after quantization~\cite{liu2022pd}.

The following three methods combine quantization and distillation techniques and make improvements based on this foundation.
Q-DETR~\cite{xu2023q} builds upon the common detection loss by incorporating Distribution Rectification Distillation loss.
This loss function aims to reduce the distortion of information between student and teacher queries through the distillation process.
Q-ViT~\cite{li2022qvit} introduces distribution-guided distillation to provide distribution-aware optimization direction and handle appropriate distilled parameters.
Specifically, a new loss function has been added based on the difference of patch-based similarity pattern matrices to eliminate scale differences in addition to the original distillation loss.
ZeroQuant~\cite{yao2022zeroquant} proposes Layer-by-Layer Knowledge Distillation, which calculates the layerwise quantization loss instead of an end-to-end manner.

From another perspective, both NWQ~\cite{zheng2022leveraging} and BSQ~\cite{yang2021bsq} introduce penalty terms through regularization techniques. 
NWQ~\cite{zheng2022leveraging} utilizes Activation Regularization $L_{AR}$ to penalize quantized intermediate activations that deviate significantly from their floating-point counterparts. 
The new loss function is composed of quantization error and $L_{AR}$ together.
BSQ~\cite{yang2021bsq} introduces Bit-level group Lasso, a differentiable bit-level sparse regularizer, and uses a hyperparameter to adjust the regularization strength.

In GPUSQ-ViT~\cite{yu2023boost}, the final quantization loss function is expressed as $L_{\text{quantize}} = \sum_l \gamma_l \cdot L_{\text{quantize}, l}$. 
The weight factor $\gamma_l$ is inversely proportional to the feature distillation loss. Layers with smaller feature distillation losses are assigned larger weight factors.
Evol-Q~\cite{frumkin2023jumping} effectively navigates through the non-smooth loss landscape encountered during the quantization of Vision Transformers by leveraging evolutionary search. 
Furthermore, it introduces the infoNCE loss function to evaluate the candidate quantization scales generated by the evolutionary search. 
In this manner, Evol-Q not only identifies superior quantization scales but also mitigates overfitting issues on small calibration datasets.

\subsubsection{Other Enlightening Mechanisms}\label{Other Enlightening Update Mechanisms}

In QuantSR~\cite{qin2023quantsr}, the Redistribution-driven Learnable Quantizer aims to introduce a nonlinear transformation function $\phi (x)$ to enrich gradient information during back propagation. 
This approach makes the quantized representation more diverse and better retains information.

PEQA~\cite{kim2023memory} is a method that only finetunes the quantization scales of quantized LLMs, while keeping the integer matrices unchanged for each downstream task, which can reduce memory consumption.
To avoid the large quantization errors caused by the delayed updates of the EMA estimator, which can lead to inappropriate quantization step sizes, BQ~\cite{bai2021batchquant} utilizes the statistical data of the current batch to estimate the extreme values of the activation values.
It also introduces multiplicative residual $\gamma$ and additive residual $\beta$ to dynamically adjust $s$ and $z$, in order to better adapt to the distribution of activation values.
QERA\cite{zhang2025qera} proposes minimizing the layer output error and derives the exact and approximate solutions through a series of matrix transformations.

\subsection{Mixed Precision}\label{Mixed-Precision}
The quantization methods previously discussed generally quantize the whole model to a fixed bit-width. 
However, this approach presents a dilemma: On one hand, with a lower bit-width, the model compresses more but suffers from greater quantization errors; 
On the other hand, with a higher bit-width, performance loss is minimized, but the acceleration benefits are less pronounced.
Researchers have already suggested that global bit-width allocation may be suboptimal for quantization, as different layers contribute differently to the model's quantization and impact inference time and task performance variably. 

\begin{figure*}
\centering
\includegraphics[width=\linewidth]{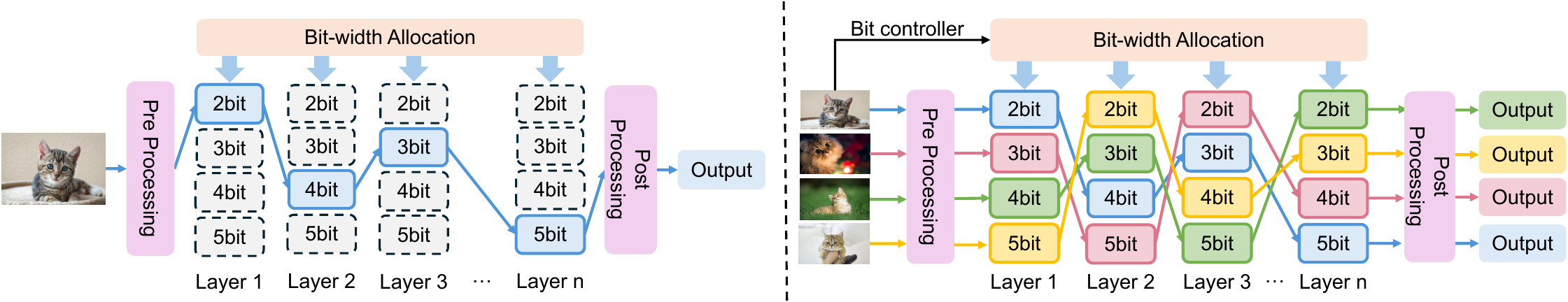}
\vspace{-2mm}
\caption{Two paradigms of mixed precision. The left one is input irrelevant, and the bit-width is allocated during the calibration process. The right one is input relevant, where a bit controller processes the image first and then dynamically decides the bit-width allocation.}
\vspace{-2mm}
\label{fig:mixed-precision}
\end{figure*}

Thus, mixed precision emerges as a bold attempt. 
For instance, HAQ~\cite{wang2019haqhardwareawareautomatedquantization} assigns more bits to important layers or channels and fewer bits to less important ones, effectively balancing accuracy and cost.
The key to mixed precision quantization is to find the optimal bit-width allocation for each layer efficiently due to the exponentially increasing searching space.
There are mainly two kinds of methods, input-static and input-dynamic, as shown in Fig.~\ref{fig:mixed-precision}.
Next, several mainstream research directions related will be summarized.

\subsubsection{Rough Allocation}\label{Rough Bit-width Allocation}
In this part, the matrix is partitioned into distinct segments based on different criteria, and each segment is allocated a different bit-width~\cite{lee2023owq,dong2019hawq,shen2019qbert,dong2020hawq,liu2021post,kim2024squeezellm,sun2022film,mills2024qua,yvinec2023rex,liu2023oscillation,shang2023ptqdm,yao2022zeroquant,chen2023data,kryzhanovskiy2021qpp,cui2024cherry}.

Several researchers carried forward the ideas of HAQ, focusing on the importance and sensitivity of weights.
OWQ~\cite{lee2024owq}, HAWQ~\cite{dong2019hawq}, CherryQ~\cite{cui2024cherry}, and Qbert~\cite{shen2020q} all analyze the Hessian matrix to identify sensitive weight columns, storing them with high precision while assigning lower precision to the rest.
However, HAWQ focuses on the top Hessian eigenvalues and ignores the rest of the Hessian spectrum.
On this basis, HAWQ-v2~\cite{dong2020hawq} uses the average Hessian trajectory to calculate the importance of the layer.
~\cite{liu2021post} uses a larger number of low-bit vectors for important channels and fewer for insensitive channels.
SqueezeLLM~\cite{kim2024squeezellm} leverages Fisher information matrix to analyze the distribution and importance of weights, and decomposes them into dense and sparse parts.
The sparse part, which contains sensitive weights and outliers, is stored with full precision.
FILM-QNN~\cite{sun2022film} calculates the quantization error of each filter.
The top 5\% of the weights are quantized using 8 bits, the remaining weights are quantized using 4 bits, and all activations are quantized using 5 bits.
For DMs, Qua$^2$SeDiMo~\cite{mills2024qua} creates a search space for the denoiser, selects the optimal bit precision and quantization methods. 
It represents the denoiser architecture as a Directed Acyclic Graph, encodes the node features, and uses GNN to learn the contribution of the operation layers to the performance. 
At the same time, it uses the scores of the root nodes of sub-graphs to represent sensitivity and constructs quantization configurations. 

Another set of studies is dedicated to identifying outliers and quantifying them with higher precision.
For example, REx~\cite{yvinec2023rex} measures the relative importance of neurons within a layer by their weight norms. 
By expanding the residual quantization error, it allows for quantization at varying bit-widths, identifying outliers in weights that can be quantized as binary values, while the rest are quantized with more bits.
OFQ~\cite{liu2023oscillation}, based on maximum information entropy theory, retains the highest entropy by uniformly distributing weight quantization, ignoring outliers in scaled weights, ensuring equal contribution in updates.

Other researchers assign varying bit-widths to cater to their specific quantization strategies.
For example, PTQ4DM~\cite{shang2023ptqdm} assigns lower bit-widths to the initial denoising steps for faster processing, ensuring high SNR and model performance in later steps for high-quality sample generation.
ZeroQuant~\cite{yao2022zeroquant} quantizes weights to INT4 in fully connected modules and maintains INT8 in attention modules.
DF-MPC~\cite{chen2023data} achieves inter-layer compensation by applying ultra-low precision to low-layer weights, using higher-precision quantization for high-layer weights, and introducing compensation coefficients to linearly reconstruct the quantization errors.
QPP~\cite{kryzhanovskiy2021qpp} integrates into dynamic quantization methods, such as Dense+Sparse quantization, where a set percentage of activations remain unquantized.

\subsubsection{Adaptive Allocation}\label{Adaptive Bit-width Allocation}
Adaptive bit-width allocation allows finer granularity, where different layers or weight groups within a network are allocated different bit-widths.
Therefore, the allocation of bit-width can be tailored according to each layer's contribution to the loss function.

Efforts have been dedicated to the design of loss functions~\cite{shin2023nipqnoiseproxybasedintegrated,chikin2022data,zhou2021optimizing}, which usually combine task-specific loss and quantization parameter loss.
For example, NIPQ~\cite{shin2023nipqnoiseproxybasedintegrated} introduces a loss function that incorporates memory consumption and computational cost, automatically adjusting the bit-width of each layer by minimizing this cost loss function.
PNMQ~\cite{chikin2022data} calculates the compression ratio, which increases monotonically with threshold $L_{opt}$, stopping iterations when the user-specified compression ratio is reached.

Some researchers have presented innovative training strategies~\cite{yang2020fracbits,yang2021bsq,huang2022sdq,shomron2021post,jin2020adabits,kim2024metamix,zhu2023rm,liu2022instance,habi2020hmqhardwarefriendlymixed,qu2020adaptive,zhang2024reshape,zhang2021differentiable}.
FracBits~\cite{yang2020fracbits} sets the bit-width of each layer or kernel as a linear combination of its two adjacent integer bit-widths, making the bit-width a continuously learnable parameter.
BSQ~\cite{yang2021bsq} proposes a gradient-based training algorithm for quantized DNN models, treating each bit of quantized weights as an independent trainable variable, which enables optimization with STE.
SDQ~\cite{huang2022sdq} introduces differentiable bit-width parameters that act as probability factors in stochastic quantization, allowing the weights to be quantized for both the current and the next lower bit-widths.
In SPARQ~\cite{shomron2021post}, it first selects the most significant four consecutive bits for quantization, then decides whether to reduce bit-width based on the sparsity of activation values.
Adabits~\cite{jin2020adabits} adopts the joint training method and uses switchable clipping level technology, configuring independent clipping levels for each bit-width.
MetaMix~\cite{kim2024metamix} independently trains the model for each candidate bit-width and calculates the average loss to update the weights. 
On this basis, it learns the bit-width probability parameters for each layer to select the optimal bit-width combination.
A2Q~\cite{zhu2023rm} leverages the topological structure of GNNs to automatically learn and allocate the appropriate bit-widths and step sizes.

DQNet~\cite{liu2022instance} uses a bit-controller to output the probability of different bit-widths.
Both~\cite{liu2022instance} and HMQ~\cite{habi2020hmqhardwarefriendlymixed} use the Gumbel-Softmax estimator to search simultaneously for both the bit-width and the threshold of each quantizer.
ALQ~\cite{qu2020adaptive} removes unimportant weights in each round, which affects the number of binary bases and thus allows more important weights to have a greater bit-width.
RAOQ~\cite{zhang2024reshape} uniformly samples different ADC bit precisions from set B.
Then the information associated with the diverse ADC bit precisions is accumulated in the backward pass through their respective gradient representations, allowing for more optimized updates to the model parameters.
DDQ~\cite{zhang2021differentiable} treats the quantization levels $q$ as trainable variables and directly optimizes the values of $q$ to adapt to the weight distribution. 
Then, it controls the merging method of the quantization levels through the block size of the block diagonal matrix $U$ to achieve dynamic bit-width adjustment.
FedMPQ~\cite{chen2024mixed} introduces a group Lasso regularization term into the objective function of local training to promote sparsity in the model parameters and judges the precision requirements of each layer based on the degree of parameter sparsity. 
Meanwhile, a threshold is set to determine whether to prune the most significant bits of the model parameters.

\subsubsection{Search Algorithm}\label{Optimized Mixed Precision Search Algorithm}
Mixed precision quantization, in the worst-case scenario, is a finite search problem. 
However, with the increasing depth of deep neural networks, brute-force search becomes impractical.
The following researches aim to reduce the search space with more constraints~\cite{yao2021hawqv3,hubara2021accurate,ma2023ompq,li2021brecq,tang2022mixed,cai2020zeroq,koryakovskiy2023one,dong2020hawq,bai2021batchquant,liu2021posttrainingquantizationvisiontransformer}.

Integer Linear Programming is a single-objective optimization method that aims to maximize or minimize a linear objective function under a set of linear constraints. 
HAWQ-V3~\cite{yao2021hawqv3}, AdaQuant~\cite{hubara2021accurate}, OMPQ~\cite{ma2023ompq}, BRECQ~\cite{li2021brecq}, and LIMPQ~\cite{tang2022mixed} calculate bit precision by solving ILP problems, balancing the disturbance of the model with other constraints.
It is worth mentioning that~\cite{yao2021hawqv3} introduces constraints directly related to hardware resources, effectively adapting to different neural network architectures and hardware architectures, reducing the inference latency and energy consumption of quantized neural network models.
BRECQ~\cite{li2021brecq} argue that the loss measurement should contain both diagonal loss and off-diagonal loss, and reduce the search space by considering only 2-bit permutations.
Moreover, it employs a genetic algorithm~\cite{whitley1994genetic} to search for the optimal bit-width allocation with a hardware performance threshold.
OMPQ~\cite{ma2023ompq} observes a positive correlation between ORM and quantization accuracy, optimizing model orthogonality through linear problems to derive the optimal bit-widths.

Pareto frontier optimization is a multi-objective optimization method that aims to find a set of balanced solutions among multiple objectives that cannot be surpassed by other solutions in all objectives at the same time. 
In more detail, for a target size of $S$, we aim to find the minimum overall sensitivity from each bit precision allocation that meets the size constraints.
The bit-width is chosen to minimize the overall sensitivity.
ZeroQ~\cite{cai2020zeroq}, One-Shot MPS~\cite{koryakovskiy2023one}, HAWQ-v2~\cite{dong2020hawq}, MultiQuant~\cite{xu2022multiquant}, QFA~\cite{bai2021batchquant}, and~\cite{liu2021posttrainingquantizationvisiontransformer} all introduce Pareto frontier-based methods to automatically determine precision bit settings for all layers without.

For instance, QFA~\cite{bai2021batchquant} employs a multi-objective evolutionary search to optimize multiple competing goals and generate effective subnets. It utilizes the NSGA II algorithm, which simultaneously provides a Pareto population, allowing quick access to the Pareto front of a trained supernet.

\subsubsection{Other Enlightening Mixed Precision Methods}\label{other}

In addition to standard approaches, several alternative search strategies have been proposed to optimize quantization configurations. 
Tang \etal~\cite{tang2024retraining} employ a greedy bidirectional search strategy that evaluates the trade-offs between increasing or decreasing bit-widths in terms of model performance and computational cost, thereby identifying an optimal allocation. 
QBR~\cite{sun2022entropy} and CLAMP-ViT~\cite{ramachandran2024clamp} adopt layer-wise evolutionary algorithms to jointly optimize architectural design and quantization precision.
ADRL~\cite{ning2021simple} integrates reinforcement learning with two types of quantization indicators. For the profiling-based indicator, at each time step $t$, the action generator produces $k$ candidate actions for layer $l$.
These are then evaluated using a validation dataset, and the candidate yielding the highest accuracy is selected. For the distance-based indicator, two metrics—L2 norm and KL divergence—are explored to determine appropriate bit-widths.
FDDA~\cite{zhong2022finegraineddatadistributionalignment} reformulates the neural architecture search (NAS) problem as a hyperparameter optimization task and utilizes Bayesian optimization to discover the optimal bit-width configuration.

\subsection{Redistribution}\label{Redistribution}
During the quantization process, adjusting the distribution of weights and activations can effectively mitigate the adverse effects of outliers in quantization. 
The essence of redistribution lies in flattening the distribution of weights and activations, thereby reducing the loss of precision during quantization. 
Redistribution also helps to minimize the accumulation of errors and distributes them more evenly across different parts of the model, thereby reducing the negative impact on the model's performance.

\begin{figure}
\centering
\includegraphics[width=\linewidth]{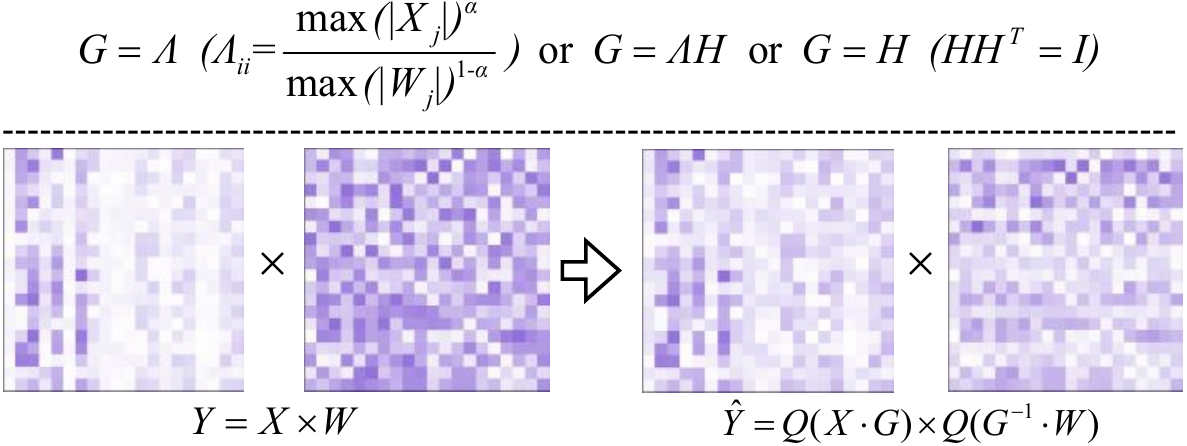}
\vspace{-2mm}
\caption{Diagram of redistribution. A widely used approach is to leverage an invertible matrix to smooth the outliers and transfer the quantization difficulty from activation to weight. There are many forms of invertible matrices to choose from.}
\vspace{-2mm}
\label{fig:redistribution}
\end{figure}

\subsubsection{Bias}\label{Bias}
QuantSR~\cite{qin2023quantsr} employs a learnable mean shift parameter, denoted as $\tau$, which enables the network to adjust the data during the quantization process. 
This adjustment ensures that the quantization intervals are more aligned with the actual distribution of the data.
Additionally, the authors introduce a transformation function, denoted as $\phi$, to modify elements that are far from the center of the quantization intervals.
NoisyQuant~\cite{liu2023noisyquantnoisybiasenhancedposttraining} addresses the heavy-tailed distribution issue in ViT by actively altering the activation distribution through the addition of a fixed uniform noise bias. 
RAOQ~\cite{zhang2024reshape} introduces a learnable shifting parameter $\beta$.The distribution of activation values is then moved away from zero, thereby increasing the variance of the activation values.Its activation quantization is represented as:
\begin{equation}
    x_{s}=\operatorname{clip}\left(x-\beta, 0,2^{b_{x}}-1\right)-2^{b_{x}-1},
\end{equation}
where $ x $ represents the original activation value, and $ b_x $ is the bit precision of the activation value.

\subsubsection{Distribution Uniformization}\label{Distribution uniformization}
Distribution uniformization involves adjusting the distribution of weights and activation to achieve a more balanced data representation, as shown in Fig.~\ref{fig:redistribution}.
By making these distributions more uniform, we can make fuller use of quantization levels, thereby maximizing information entropy.
The benefits of this approach are manifold: 
Firstly, it enhances the model's resilience to quantization errors, making the model more robust in the face of changes in quantization step size;
Secondly, it helps to prevent the model's quantization outcomes from being dominated by the majority class in the data, thus improving the model's generalization ability.

ClimbQ~\cite{chen2022climbq} is a quantization method designed for datasets with class imbalance, aiming to improve the accuracy of minority classes during the quantization process. 
In datasets with class imbalance, the quantization outcomes tend to be dominated by the majority classes, which leads to a significant decrease in the accuracy of minority classes. 
To address this issue, ClimbQ increases the variance of minority classes and decreases the variance of majority classes through scaling, reducing the differences between class distributions. 
It then uses projection to transform data into a uniform space, achieving uniform quantization.

RAPQ~\cite{yao2022rapqrescuingaccuracypoweroftwo} employs two techniques to adjust the statistical properties of activations and weights: the activation shift method and the weight reshaping technique.
The A-shift method treats activation values as unsigned numbers for quantization.
It then performs a shift operation on the quantized unsigned activation values, moving their distribution further away from zero, which increases variance. 
After the shift, these unsigned activation values are converted back to signed numbers. 
The W-reshape technique, on the other hand, adjusts the distribution of weights by introducing a kurtosis penalty, which helps to flatten the distribution of the weights. 
This flattening process increases the variance and leads to a more uniform distribution, enhancing the quantization process.
KURE~\cite{shkolnik2020robustquantizationmodelrule} also employs this method to redistribute weights.

Q-DM~\cite{li2023qdm} observes that after each time step, there may be significant variations in the distribution of activation values, which could lead to oscillations in the activation distribution. To address this issue, QDM draws on the concept of Batch Normalization to make the activation distribution more uniform by introducing quantization-aware attention blocks. Specifically, QDM first normalizes the query and key vectors, and then normalizes the values calculated after the softmax function is applied. This process smooths out the distribution of activation values.

\subsubsection{Outlier Processing}\label{Outlier Processing}

Large Language Models (LLMs) often have a significant number of outliers in their weights and activation values, particularly in the activation values where outlier channels frequently occur. 
This makes the quantization of LLMs challenging. 
Consequently, many researches focus on processing weights and activations to achieve a more uniform and smooth distribution, thereby reducing the impact of outliers on quantization errors and improving performance.

SmoothQuant~\cite{xiao2024smoothquantaccurateefficientposttraining} addresses this by scaling activations and weights channel by channel.
The authors divide each element in each column of the activation matrix by a smoothing factor and multiply each element in each row of the weight matrix by $ s_i $ to maintain the mathematical equivalence of the linear layer.
SmoothQuant alters the distribution of weights and activations, transferring the quantization difficulty from activations to weights, balancing outliers between the two. 
Additionally, ~\cite{wei2023outliersuppressionaccuratequantization} introduced channel-wise offsetting, $ (X-z) \cdot \text{diag}(c)^{-1} $, to further mitigate the impact of outliers, while Omniquant~\cite{shao2024omniquantomnidirectionallycalibratedquantization} treated $ \text{diag}(c) $ and $ z $ as learnable parameters.
FQ-Vit~\cite{lin2023fqvitposttrainingquantizationfully} assigns a unique scaling factor $2^\alpha$ to each channel of the LayerNorm layer, thereby readjusting the quantization range for each channel.

However, SmoothQuant does not distribute outliers across non-outlier channels. 
Research has found that Hadamard matrices $ H \in \{+1, -1\}^{n \times n} $ are particularly helpful in smoothing outliers in activations~\cite{ashkboos2024quarotoutlierfree4bitinference,lin2024qserve,xi2023trainingtransformers4bitintegers,tseng2024quip,feng2024mpq}. 
Unlike channel-wise scaling, which only adjusts diagonal elements from a matrix multiplication perspective, Hadamard transforms rotate the channels of activations and weights, redistributing outliers across all channels to effectively eliminate them. 
Multiplying by a random Hadamard matrix ensures the incoherence of weights and the Hessian matrix, reducing outliers. Due to the orthogonality of Hadamard matrices (\ie $ H^T H = I $), the following equivalence holds: $ Y = X W^T = (X H) (H^T W^T) $. The transformed weights $ W H W^T $ can also be pre-processed offline to reduce runtime overhead.

DuQuant~\cite{lin2024duquantdistributingoutliersdual} targets the channel dimensions containing outliers, using a rotation matrix to redistribute them to neighboring channels via block-wise rotation, thus minimizing their effect. 
It also introduces a zigzag permutation to evenly distribute outliers across blocks, reducing inter-block variance. Lastly, an additional rotation transformation is applied to smooth the activation landscape.

~\cite{Nie2022RedistributionOW} eliminates outliers by clamping weights within a range $[-r_x, r_x]$ and then adding a bias term $ b $ to compensate for the effects of the clamping.
MagR~\cite{zhang2024magrweightmagnitudereduction} adjusts the weights during the pre-processing stage.
Specifically, they solve an optimization problem defined as:
\begin{equation}
    \min_{W} \frac{1}{2} \|XW - X\hat{W}\|_F^2 + \alpha \|W\|_{\infty},
\end{equation}
where $\| \cdot \|_F$ denotes the Frobenius norm, which measures the error of the matrix, and $\alpha$ is the regularization parameter that balances the trade-off between the fidelity of the layer output and the magnitude of the weights.

\subsubsection{Rounding}\label{Rounding}
Rounding strategies determine how floating-point numbers are rounded to the nearest quantization levels. Optimized rounding strategies can adjust the distribution of quantized values, allowing the quantized model to more accurately represent the behavior of the original model, thereby enhancing the model's performance.

AdaRound~\cite{nagel2020downadaptiveroundingposttraining} introduced a theoretically sound and computationally efficient weight rounding method. 
During the quantization, AdaRound does not use the traditional round-to-nearest approach.
Instead, it adaptively determines whether to round the floating-point value to the nearest positive or negative fixed-point value~\cite{nagel2020downadaptiveroundingposttraining}. Several methods, including BRECQ and QDrop, build upon the AdaRound technique.

AdaRound requires that the quantized weights must be within the $\pm 1$ range of their nearest rounding values.
AdaQuant~\cite{hubara2020improvingposttrainingneural} relaxes this constraint by considering non-adjacent fixed points in the rounding process. 
This is achieved by optimizing the weights and quantization parameters of each layer separately over the calibration set, with the goal of minimizing the Mean Squared Error between the layer's original and quantized outputs. 
Additionally, unlike AdaRound, AdaQuant is used not only for the optimal quantization of weights but also for the quantization of activations.
In the papers~\cite{yang2020searchinglowbitweightsquantized, wang2022leveraging}, researchers have utilized a temperature parameter $\tau $ to precisely control the probability distribution of weights in quantized neural networks. 
Researchers define a probability distribution to calculate the likelihood of weights being quantized to various discrete values.
To identify the discrete value with the highest probability, researchers gradually reduce $\tau$.
As $\tau$ approaches zero, the network can deterministically select the discrete value with the highest probability as the quantized output.

\subsection{Data-Free Quantization}\label{sec:data-free-quantization}
\begin{figure}
\centering
\includegraphics[width=\linewidth]{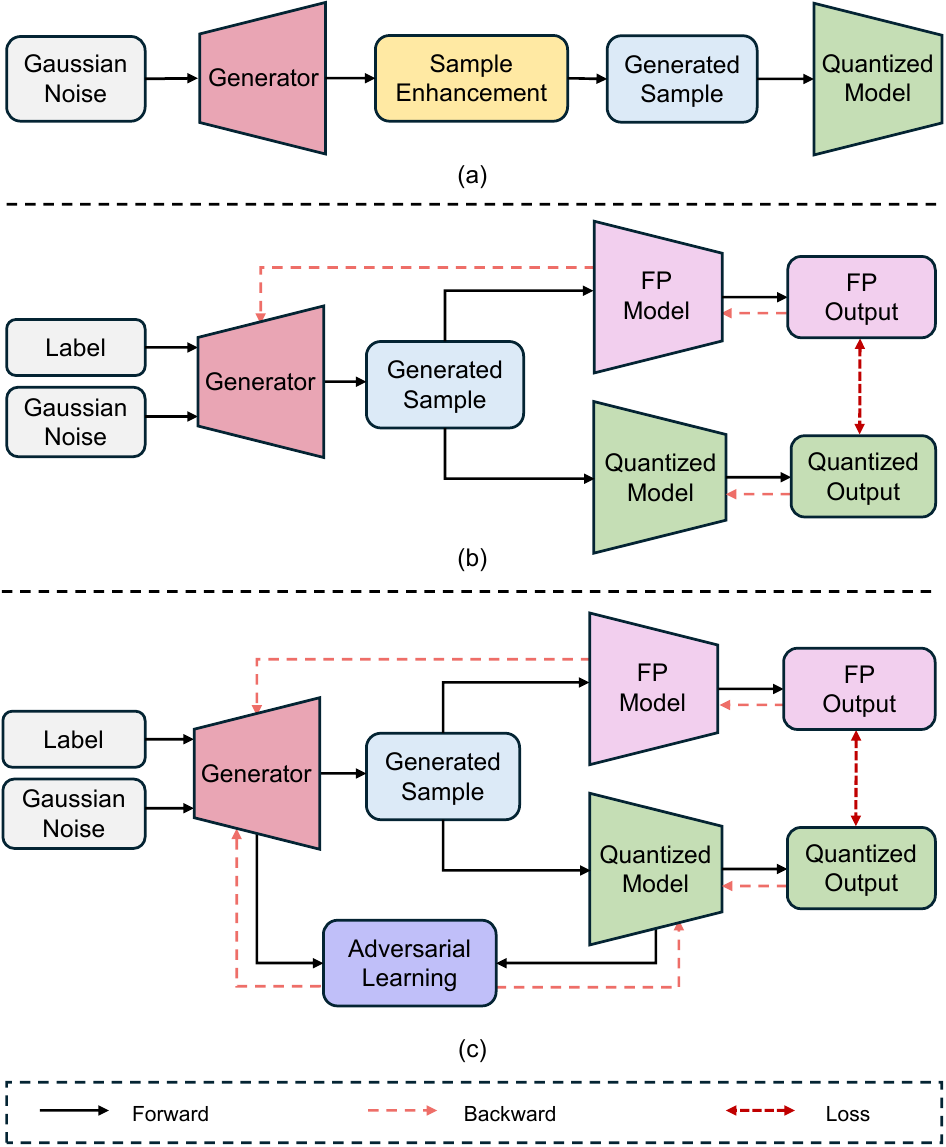}
\vspace{-2mm}
\caption{Three data-free quantization paradigms. (a) The generator takes Gaussian noise as input to obtain raw samples, which will be enhanced with information from the FP model or the quantized model. (b) The FP model is engaged in the loss function, and the gradient will be back-propagated to the generator for optimization. (c) There's adversarial learning between the generator and the quantized model. The quantized model function is similar to the discriminator in GAN. }
\label{fig:data-free-quantization}
\vspace{-2mm}
\end{figure}

As previously discussed, both QAT and PTQ necessitate calibration data for calibration or fine-tuning during the network quantization process. 
However, in numerous practical applications, access to training data may be restricted due to privacy or confidentiality concerns. 
In such cases, these methods are no longer applicable.
Thus, data-free quantization (DFQ) emerges as a highly valuable alternative, offering a practical solution to the challenges of model quantization and acceleration without the need for original training datasets.
DFQ, in essence, proposes to carry out quantization without accessing the training data of full-precision models. 
Related research can be divided into several branches, shown in Fig.~\ref{fig:data-free-quantization}.

\subsubsection{Truly Data-Free Methods}\label{sec:truly-data-free-methods}
These truly data-free methods~\cite{nagel2019data, guo2022squant, yvinec2023rex, chikin2022data} use effective strategies to directly quantize the model's weights and activations without any calibration data.

For instance, DFQ~\cite{nagel2019data} quantizes through weight equalization and bias correction, instead of fine-tuning on the entire dataset. 
SQuant~\cite{guo2022squant} employs diagonal Hessian approximation on a layer-by-layer basis.
REx~\cite{yvinec2023rex} quantizes residual errors during the quantization process.
It is anticipated that the predictions of the quantized model will more closely approximate those of the original model as the quantization error diminishes. 
PNMQ~\cite{chikin2022data} utilizes a non-uniform quantization grid defined by scale factor $s$ and parameter $p$, optimizing these parameters by minimizing the norm of the quantization error. 
UDFC~\cite{bai2023unified} suggests that the partial loss of information in a damaged channel can be compensated by combining other channels. 
It formulates a layer-wise reconstruction approach and reduces the reconstruction error to recover information lost during compression.

\subsubsection{Accurate and Diverse Generation}\label{sec:accurate-and-diverse-generation}
A straightforward concept is to sample random inputs from a distribution and use them to calibrate the model.
Unfortunately, random inputs are significantly divergent from the real data distribution, resulting in substantial performance degradation. 
Consequently, the accurate and diverse calibration data generation for model quantization presents a formidable challenge.
It's worth noting that the samples from an accurate distribution are not always meaningful.
In real-world practice, the samples are mostly meaningless but contain crucial patterns.

Therefore, synthetic samples are often derived through noise optimization, where the noise distribution is optimized to align with the statistics of real data distributions.
Many studies have made contributions to generate accurate and diverse data~\cite{cai2020zeroq, xu2020generativelowbitwidthdatafree, zhang2021diversifyingsamplegenerationaccurate, zhong2022intraqlearningsyntheticimages, li2023hardsamplematterslot, chen2023texq, li2023normtweakinghighperformancelowbit, li2022psaqvit, yvinec2023powerquantautomorphismsearchnonuniform, xu2022multiquant, kimsynq, choi2021qimera,bulat2024qbb,shang2023causal,fan2024data,liu2021zeroshotadversarialquantization,Qian_Wang_Hong_Wang_2023,qian2023adaptive}.

ZeroQ~\cite{cai2020zeroq} is one such method that uses the statistical information of BN layers to constrain and backpropagate the input data.
Based on ZeroQ, GDFQ~\cite{xu2020generativelowbitwidthdatafree} generates samples guided not only by batch normalization statistics but also incorporating additional category label information, ensuring that the generated data retains classification boundary knowledge and data distribution.
Targeting Transformers, PSAQ-ViT~\cite{li2022psaqvit} makes use of the self-attention module instead of the BN layers for sample generation and employs differential entropy to seek data diversity.
Norm Tweaking~\cite{li2023normtweakinghighperformancelowbit}, QBB~\cite{bulat2024qbb} and PowerQuant~\cite{yvinec2023powerquantautomorphismsearchnonuniform} all incorporate distillation concepts. 
They design frameworks to minimize the distribution difference between the quantized model and the original floating-point model.
Causal-DFQ~\cite{shang2023causal} constructs a causal graph to model the data generation process.
Through Do-Calculus operations, it enforces the model outputs to remain invariant when the style variable changes, eliminating the influence of irrelevant factors and generating fake images conditioned on style and content.

Researchers also notice that the generated samples tend to cluster around class centroids in image classification, leading to homogenization.
Qimera~\cite{choi2021qimera} addresses this issue with superposed latent embeddings and soft labels.
With these techniques, boundary-supporting samples are synthesized to enable the quantized model to learn more precise decision boundaries.
Similarly, MultiQuant~\cite{xu2022multiquant} and SYNQ~\cite{kimsynq} also use soft labels to guide the training. 
Besides, SYNQ tries to reduce noise in synthetic data through a low-pass filter and aligns the class activation map to ensure correct prediction regions.
~\cite{fan2024data} employs self-entropy to measure the reliability of synthetic data, categorizing it into high-reliable and low-reliable datasets.
It guides the quantized network by designing pseudo-labels and loss functions.

Recognizing the homogenization at the sample level caused by applying the same constraints to all data samples, DSG~\cite{zhang2021diversifyingsamplegenerationaccurate} assigns a unique layer-wise loss to each sample, forcing different samples to optimize the statistics of different BN layers and enhancing diversity among samples. 
There is one observation that NNs extract class-independent low-level features in shallow layers and class-related high-level features in deep layers~\cite{chen2023texq}. 
TexQ~\cite{chen2023texq} constructs loss functions based on this characteristic.
This allows rich texture features to be preserved in shallow layers and ensures alignment of class-related semantic features in deep layers, thereby enhancing sample diversity.
Intraq~\cite{zhong2022intraqlearningsyntheticimages} observes the importance of intra-class heterogeneity and tries to retain this property.
Unlike IntraQ, HAST~\cite{li2023hardsamplematterslot} places greater emphasis on increasing sample difficulty, making synthetic samples more challenging for the quantized model to fit.

\subsubsection{Adversarial-Based Generation}\label{sec:adversarial-based-generation}
Since the performance depends heavily on synthetic samples, some researchers focus on improving the performance by adopting adversarial training.
Adversarial learning is a learning paradigm in which two or more agents interact under the backdrop of a zero-sum game, each with its own objective, which are typically in opposition to each other.

ZAQ~\cite{liu2021zeroshotadversarialquantization} employs the two-level discrepancy modeling.
In the discrepancy estimation phase, the generator $G$ aims to maximize the discrepancy between the full-precision model $P$ and the quantized model $Q$.
In the knowledge transfer phase, $Q$ aims to minimize the discrepancy.
AdaSG~\cite{Qian_Wang_Hong_Wang_2023} redefines the DFQ problem as a zero-sum game between $G$ and $Q$ and puts forward the definition of sample adaptability, which measures the degree to which generated samples are beneficial or harmful to the learning process.
Building upon this concept, AdaDFQ~\cite{qian2023adaptive} optimizes the margin between disagreement and agreement samples to dynamically adjust the adaptability, thereby addressing overfitting issues.

\subsection{Advanced Format}\label{Advanced format}
The conventional integer formats of quantization are often characterized by a fixed dynamic range.
This may not be suitable for the broad distribution of weights and activation values in DNN, thereby limiting the models' expressive capabilities. 
Furthermore, traditional integer quantization needs extensive manual tuning and fine-tuning to accommodate various models and tasks, thereby increasing the complexity of development and deployment.
To address these inefficiencies, research has focused on identifying encoding formats that offer higher precision, a wider dynamic range, or are more compatible with hardware.
The ultimate goal is to enhance the performance, energy efficiency, and memory requirements of model inference.

\subsubsection{Float-Based}\label{FP-based}

The traditional floating-point (FP) format can be written as:
\begin{align}
\text{Real value}=\operatorname{sign} \times 2^{\text {exponent value}} \times \text{mantissa value}.
\end{align}
FP16~\cite{kahan1996ieee} uses a 5-bit exponent and 10-bit mantissa (5E10M).
8E7M and 8E10M are used by BF16~\cite{bf16} and TF32~\cite{chmiel2020neuralgradientsnearlognormalimproved} respectively.
FP allows the coverage of a broad dynamic range with a relatively small number of bits. 
Researchers are dedicated to exploring the optimal allocation of the bit-width between the mantissa and the exponent for an FP number of N bits.
TF32~\cite{chmiel2020neuralgradientsnearlognormalimproved} defines the number of bits for the mantissa as $n_{1}$ and for the exponent as $n_{2}$.
Since one bit is reserved for the sign of the number, $n_{1}=N-n_{2}-1$.
By solving the derivative of the error formula set to zero, the optimal values for $n_{1}$ and $n_{2}$ can be determined.

AdaptiveFloat~\cite{tambe2020algorithm} dynamically sets the tensor-wise exponent bias to match the weights' matrix and minimize quantization error.
ANT~\cite{guo2022antexploitingadaptivenumerical} dynamically allocates the number of bits for the exponent and mantissa based on the value's magnitude, and uses the position of the first occurrence of 1 to mark the boundary between the exponent and mantissa.
The authors of DFloat~\cite{zhang202570size100accuracy} observed that the entropy of the exponent part in BF16 weights is significantly lower than its allocated bit-width, indicating substantial redundancy.
Based on this insight, they apply Huffman coding to the exponent part of the BF16 weights.
For the sign and mantissa bits, they kept them unchanged.
This format allows them to compress the weights of LLMs to approximately 11 bits without any loss of precision.
MSFP~\cite{darvishrouhani2020pushing} shares an exponent across a group of values, known as a "bounding box".
The mantissa of each value is then adjusted based on the shared exponent and its own individual exponent.
Building upon MSFP, BSFP~\cite{lo2023block} introduces an enhancement that uses the sum of two sub-word vectors, each with its own scaling factor, to approximate each full-precision weight vector. 
Each subword is a low-bit integer, and each scaling factor is a low-bit floating-point number. 
This enables BSFP to capture larger weights with a subword vector that has a larger scaling factor, while a subword vector with a smaller scaling factor mitigates the remaining discrepancies. 

\subsubsection{Fixed Point}\label{Fixed Point}
Multiplication of floating-point numbers often encounters inefficiency.
To address this, F8Net~\cite{jin2022f8net} and Vs-quant~\cite{dai2021vs} employed fixed-point data type. 
A fixed-point number is characterized by its word length (WL) and fraction length (FL).
WL defines the total bit-width and FL specifies the range and resolution~\cite{tan2018digital}.
Since the optimal representation regions vary when fixed-point numbers are in different formats, the focus of F8Net~\cite{jin2022f8net} is to determine the most suitable FL.
The authors of F8Net~\cite{jin2022f8net} plotted the threshold standard deviation $\sigma$ against the corresponding optimal FL, and they observed an almost linear relationship between them. 
This observation has led to the derivation of a semi-empirical approximation formula as follows:
\begin{equation}
    \text{Signed:} \mathrm{FL}^{*} \approx\left\lfloor\log _{2} \frac{40}{\sigma}\right\rfloor, \text{Unsigned:} \mathrm{FL}^{*} \approx\left\lfloor\log _{2} \frac{70}{\sigma}\right\rfloor.
\end{equation}
 
\subsubsection{Other Formats}\label{other formats}
Qlora~\cite{dettmers2023qloraefficientfinetuningquantized} introduces NormalFloat, a method specifically designed to optimize weights with a zero mean that follow a normal distribution.
First, Qlora estimates the $2^k + 1$ quantiles of $\mathcal{N}(0, 1)$ to derive a k-bit quantile.
These quantiles are then normalized to the $[-1, 1]$ range. 
Finally, the weights are quantized using the normalized quantiles.

LoftQ~\cite{li2023loftq} also used the NF.
BiE~\cite{zou2024bie} introduced the bi-exponent mechanism, where a shared exponent $e_n$ is used for the normal values, and another shared exponent $e_o$ is used for outliers.
These two parts are distinguished by a predefined threshold $T$.
For a vector $\mathbf{X}$ containing $N$ elements, it can be written as:
\begin{align}
    2^{e_{n} \mid e_{o}}\left[(-1)^{s_{0}} m_{0}^{\prime \prime},(-1)^{s_{1}} m_{1}^{\prime \prime}, \ldots,(-1)^{s_{N-1}} m_{N-1}^{\prime \prime}\right],
\end{align}
where $s_i$ is the sign bit, $ m_i^{\prime\prime}$ is the private mantissa, $t_i$ is the type bit, used to indicate whether the element belongs to the normal values part or the outliers part~\cite{zou2024bie}.
LUT-GEMM~\cite{park2024lutgemmquantizedmatrixmultiplication} accelerates matrix multiplication by utilizing a lookup table (LUT).
It processes quantized weights directly to avoid additional overhead and uses the LUT to reduce redundant calculation.

\subsection{Diffusion Model}\label{sec:DM}
Diffusion models, which originated from the Denoising Diffusion Probabilistic Models (DDPMs)~\cite{ho2020denoising}, have emerged as a new SOTA in deep generative modeling in image generation tasks. 

\subsubsection{Brief Introduction to DMs}
The diffusion model involves two key processes: the forward process and the reverse process, both of which are parameterized Markov chains.
During the forward process, images $x_{0}$ from the training set are subjected to $T$ iterations of noise addition, resulting in $x_{T}$, which conforms to a standard normal distribution.
In the reverse process, a denoising procedure is employed to transform a noise image sampled from the standard normal distribution back into the original image, achieving the goal of image generation.

The iterative denoising procedure brings astonishing performance and unavoidably slow inference speed, gradually becoming a major concern.
DDIM~\cite{song2022denoisingdiffusionimplicitmodels} addresses this issue through mathematical deduction, bypassing the constraints of the Markov chain process.
Without altering the forward noise addition, DDIM enables sampling across multiple steps, which substantially accelerates the sampling speed.

\begin{figure*}
\centering
\includegraphics[width=0.95\linewidth]{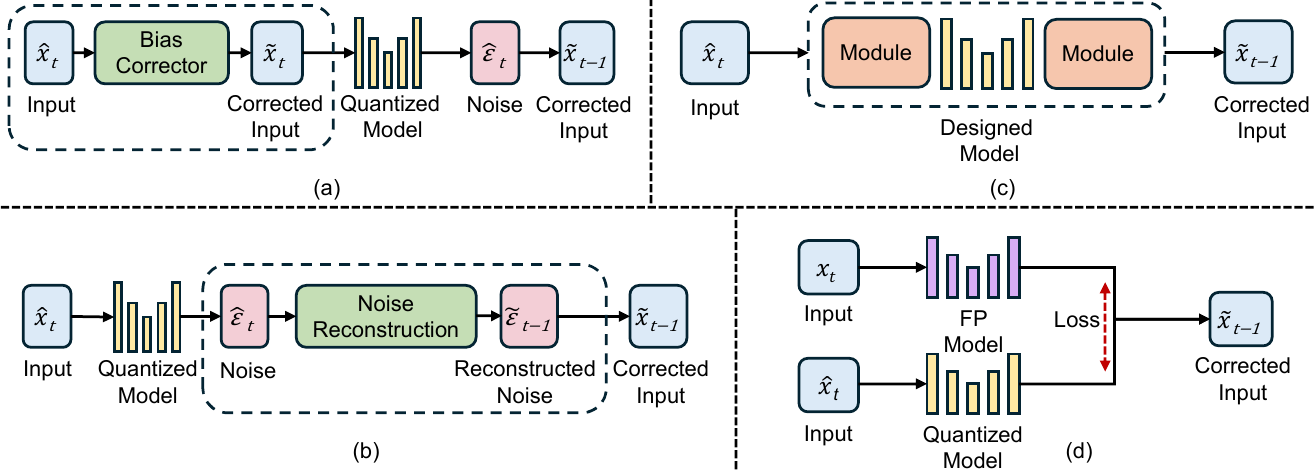}
\vspace{-2mm}
\caption{Four schematic diagrams of diffusion model quantization. There are relevant studies on the input, model structure, output and calibration methods of the diffusion model to improve the performance of the quantized diffusion models.}
\label{fig:diffusion-model}
\vspace{-2mm}
\end{figure*}

However, diffusion models face several common issues regarding quantization:
1) The output distribution of the activation layers in diffusion models changes with time steps. 
2) Quantization errors accumulate at each time step, severely impacting the model's accuracy.
Several studies have adapted their methods to the unique multi-time-step processes and model architectures of diffusion models, with the goal of accelerating the generation process by compressing noise estimation networks, shown in Fig.~\ref{fig:diffusion-model}.

\subsubsection{Calibration Set}\label{sec:Calibration}
To mitigate the accumulated error at each time step, researchers first focus on the calibration process~\cite{shang2023ptqdm, li2023qdiffusion, Huang_2024_CVPR, wang2024towards}.
PTQ4DM~\cite{shang2023ptqdm} introduces normally distributed time-step calibration, which samples time steps from a skewed normal distribution, generates calibration samples corresponding to these time steps through the denoising process.
Recognizing the similarity in activation distribution across consecutive time steps, Q-Diffusion~\cite{li2023qdiffusion} proposes a step-aware uniform sampling strategy to create a small calibration set that captures the distribution across all time steps. 
Additionally, TFMQ-DM~\cite{Huang_2024_CVPR} observes that the maximum time step for denoising is a finite positive integer.
Therefore, it utilizes finite set calibration for the discrete set of time features and activations during the generation process.
APQ-DM~\cite{wang2024towards} selects the best time steps for generating calibration images to enhance their representativeness and information content.
Various calibration metrics are leveraged by these methods, including L1 distance, cosine distance, KL divergence, and MSE.

Another way is to accept and compensate for the quantization error.
PCR~\cite{tang2024post} uses a progressive calibration strategy to generate calibration data with quantized models, ensuring that the calibration data includes the impact of quantization errors from preceding steps.
D$^{2}$-DPM~\cite{zeng2025d} discovers that quantization noise follows a Gaussian distribution at each time step.
So it quantifies the joint Gaussian model of the quantized output and quantization noise to make use of their mean and variance for better noise reduction.

\subsubsection{Activation Distribution}\label{Challenges Concerning Activation Distribution}
Researchers also focus on the varying activation distribution at each time step~\cite{li2023qdiffusion, li2023qdm, wang2024towards, zheng2024bidm}.
Q-Diffusion~\cite{li2023qdiffusion} observed that the to-be-concatenated features from skip-connection and up-sampling are quite different in the UNet architecture.
Therefore, it quantifies these two kinds of features with different quantizers.
Q-DM~\cite{li2023qdm} addresses the oscillation issue in activation distribution by smoothing activation values to make them less sensitive to random sampling time steps and time step-aware quantization during training.
TDQ~\cite{so2023temporal} employs geometric Fourier frequency encoding to transform the time step into a high-dimensional feature vector.
It then maps the encoded features to quantization intervals through an MLP to adapt to the activation distributions of different time steps.
And APQ-DM~\cite{wang2024towards} designs quantization functions for each group based on the changing characteristics of activation distribution across all time steps.
BiDM~\cite{zheng2024bidm} and TuneQDM~\cite{ryu2024memory} both observed that applying a fixed scaling factor across all time steps fails to adapt to their diversity and leads to significant distortion of the activation range.
Thus, TuneQDM partitions the time steps into $N$ intervals and independently trains a dedicated set of scale parameters for each interval.
And BiDM employs a dynamical binary quantizer, allowing the model to adapt based on input ranges with a learnable scaling factor $k$.
Additionally, an inter-timestep information enhancement connection is established to enrich the expression of the current time step using features from the previous one.

\subsubsection{Quantization Error}\label{Quantization error}
The third branch of work~\cite{he2023ptqd, li2023qdm, sui2024bitsfusion, zhao2024mixdq, chen2024stepbaq, yao2024timestep} notices significant quantization errors caused by the multi-step denoising process, which is mainly addressed during the forward process.
PTQD~\cite{he2023ptqd} proposes decomposing quantization noise into two parts, the correlated part and the uncorrelated residual part. 
PTQD corrects the correlated part by estimating the correlation coefficient and absorbs the uncorrelated part into Gaussian noise~\cite{he2023ptqd}.

Q-DM~\cite{li2023qdm} incorporates a full-precision DM to force the quantized model to replicate the noise estimation capabilities of the full-precision model. 
BitsFusion~\cite{sui2024bitsfusion} designed a sampling strategy to utilize the Beta distribution in order to sample more time steps that display maximum error.
MixDQ~\cite{zhao2024mixdq} reveals that the initial token in sequential data exhibits significantly larger magnitudes compared to subsequent tokens. 
To mitigate this outlier effect, it precomputes the full-precision output features specifically for the Beginning of Sentence (BOS) token layers and then concatenates them with the dequantized features of other tokens.
Both MPQ-DM~\cite{feng2024mpq} and SVDQuant~\cite{li2024svdqunat} build upon SmoothQuant's method of migrating outliers in activations to weights. 
MPQ-DM selects features from multiple consecutive time steps as smoothed distillation targets, thereby mitigating feature instability. 
In contrast, SVDQuant employs a low-rank branch to absorb weight outliers, minimizing the magnitude of the residuals and reducing quantization errors.

StepbaQ~\cite{chen2024stepbaq} and TAC-Diffusion~\cite{yao2024timestep} both choose to calibrate at each time step. 
StepbaQ proposes variable adjustment, which involves subtracting the mean quantization error from quantized variables and subsequently scaling them to bridge distribution shifts, thus improving the accuracy of noise prediction. 
Additionally, the method incorporates adaptive step size control, effectively reducing error accumulation through dynamic adaptation.
In TAC-Diffusion~\cite{yao2024timestep}, the NER module introduces a coefficient to achieve linear scaling for partially reconstructing the full-precision noise estimation.
Complementarily, the IBC module corrects the input bias shift between training and inference phases, reducing the exposure bias.

\subsubsection{Other DM Quantization Methods}\label{sec:other-DM}
In addition to the research directions mentioned above, a growing body of work in recent years has reported innovative discoveries.
ViDiT-Q~\cite{zhao2024vidit} incorporates static-dynamic channel balancing and metric decoupled mixed precision design to achieve the best visual generation performance.
Q-DiT~\cite{chen2024q} identifies that channel-wise quantization needs frequent intermediate rescaling operations, which fundamentally constrains the computational efficiency. 
To resolve this, it proposes adaptive group-wise quantization with automated cluster allocation guided by FID optimization.
By simplifying the OSEDiff architecture to a core UNet-VAE structure, PassionSR~\cite{zhu2024passionsr} proposes the Learnable Equivalent Transformation and Learnable Boundary Quantizer, which are designed to dynamically optimize activation distributions.
DGQ~\cite{ryu2025dgq} emphasizes that outliers are crucial for maintaining image quality and identifies that outliers occur on specific channels and pixels. It employs k-means for group-wise quantization to minimize errors while retaining outliers.

\subsection{Other}\label{sec:other}
Some quantization methods present a novel perspective that can not be simply classified into any categories we present.
In this condition, we classify these novel papers into the following four inspiring sub-categories.

One representative branch is about full-quantization.
Full quantization means converting the entire model's computational graph to use only integer operations or bit-shifting, which can skip the quantization and dequantization process.
However, in VIT, quantizing some non-linear layers, such as LayerNorm, Softmax, and GELU, can lead to serious accuracy loss, which poses a big challenge for full quantization.
I-Vit~\cite{li2023ivitintegeronlyquantizationefficient}, FQ-Vit~\cite{lin2023fqvitposttrainingquantizationfully}, Jetfire~\cite{xi2024jetfireefficientaccuratetransformer} and I-bert~\cite{kim2021ibertintegeronlybertquantization} are dedicated to solving this problem.
Besides,~\cite{lin2020fully8bitintegerinference} introduced the Scale Propagation algorithm to resolve the scale incompatibility issue of different INT8 tensors. 
It also replaced the exponential function in the standard attention mechanism and the square root function in layer normalization to make them more INT8-friendly.
Moreover, AQ-DETR~\cite{wang2024aq} is designed specially for the detection transformer, and Reg-PTQ~\cite{10654966} is applicable to various object detection models, including but not limited to DETR.
And ~\cite{tianqi2023towards} uses full Quantization to enhance the efficiency and accuracy of Winograd convolution.

The quantization methods to be presented are about the combination of quantization and other model compression techniques.
GETA~\cite{qu2025automaticjointstructuredpruning} introduces the quantization-aware dependency graph, which facilitates the construction of a pruning search space for quantization-aware DNNs. 
Additionally, it presents the quantization-aware structured sparse optimizer, which integrates a novel joint learning strategy to address the conflicts when combining pruning and quantization.
Meanwhile, there are papers employed mixed-strategy knowledge distillation to compensate for the accuracy loss~\cite{guo2024compressing,huang2022sdq,boo2021stochastic,zhu2023quantizedfeaturedistillationnetwork}.

The third branch is about stochastic quantization.
One way to leverage the stochasticity is similar to Dropout, which randomly selects only part of the model to quantize, ensuring model robustness.
Different granularities are adopted, where QDrop~\cite{wei2023qdrop} randomly quantizes the whole layer while Quant-Noise~\cite{fan2021training} prefers a sub-block of the weights.
Observing that the nodes with higher in-degree in the graph neural network are usually more important, Degree-Quant~\cite{tailor2021degree} selects these high in-degree nodes more frequently for full-precision calculations.
An alternative approach involves selecting the bit-width randomly within a mixed precision setting. 
SPEQ~\cite{boo2021stochastic} introduces the concept of stochastic bit precision, defined as:
\begin{equation}
n_{\mathrm{SPP}}^{l} = \begin{cases}
n_{A}, & \text{with probability } u, \\
n_{H}, & \text{with probability } 1 - u,
\end{cases}
\end{equation}
where $l$ represents the layer index, $n_A$ is the target bit-width, $n_H$ is bit-width greater than $n_A$, and $u$ denotes the probability~\cite{boo2021stochastic}.
SDQ~\cite{huang2022sdq} introduces differentiable bit-width parameters, representing the selection of discrete bit-widths as a set of probability factors.
These learnable parameters are updated with the loss function and gradient descent.

Besides, SDP4Bit~\cite{jia2024sdp4bit4bitcommunicationquantization}, QSDP~\cite{markov2023quantizeddistributedtraininglarge} is specially designed for sharded data parallelism.
~\cite{dettmers20228bitoptimizersblockwisequantization} uses block-wise dynamic quantization to reduce the optimizer state from 32 bits to 8 bits.
~\cite{zhao2021distributionadaptiveint8quantization} focuses on how to efficiently quantify the gradients in backpropagation.
~\cite{lee2021networkquantizationelementwisegradient} introduces element-wise gradient scaling to replace STE, while ~\cite{liu2022nonuniformtouniformquantizationaccuratequantization} proposes generalized straight-through estimator.
~\cite{chu2023makerepvgggreateragain} modify the network architecture to make it more suitable for quantization.

\section{Future}\label{sec:future}
\subsection{Application in Multimodal Tasks}  
Multimodal tasks, like text-to-image and text-to-video generation, are advancing rapidly, relying on large-scale models that integrate Vision Transformers and Large Language Models. However, their large scale demands significant computational and memory resources, which can limit scalability and real-world use, particularly in resource-constrained environments. Diffusion models, for example, require large computations for tasks like image generation. Quantization reduces model weight and activation precision, improving efficiency and enabling deployment on devices with limited resources, such as mobile phones. Future research will likely focus on refining quantization techniques to enhance performance and reduce computational costs, facilitating broader adoption.

\subsection{Combination Between Compression Methods}  
As DNNs grow in scale and complexity, model compression techniques, especially quantization, become crucial. However, quantization alone often falls short in addressing deployment and performance challenges. This has led to combining quantization with other techniques like pruning and low-rank decomposition. Pruning reduces parameters, improving memory and computational efficiency, while low-rank decomposition further compresses models by approximating weight matrices with lower-rank matrices. These hybrid methods enable more aggressive compression with minimal performance loss.

\subsection{Software-Hardware Co-Optimization}
Software-hardware co-optimization focuses on designing quantization strategies that are tailored to the constraints and capabilities of the target deployment hardware, such as NPUs, FPGAs, or edge accelerators. Rather than treating quantization as a purely algorithmic concern, this approach takes into account factors like memory bandwidth, parallelism, instruction sets, and hardware-specific operations. Future directions involve building integrated toolchains that co-design quantized neural architectures and execution plans, resulting in models that achieve optimal performance, energy efficiency, and latency when deployed on specific devices.

\subsection{Task- and Model-Specific Quantization}
Task- and model-specific quantization adapts the strategy to the neural network architecture and application needs. Different tasks, like image classification or language modeling, vary in sensitivity to quantization noise, and model components such as attention mechanisms or convolution blocks also respond differently. By tailoring quantization to both the model's structure and task performance requirements, this approach improves deployment efficiency while maintaining accuracy and robustness, especially in domain-specific edge scenarios.

\section*{Acknowledgments}
We sincerely appreciate Dehui Wang and Jiachen Qin for participating in the collection of articles in the early stage.

\bibliographystyle{IEEEtran}
\bibliography{main}

\vspace{ -15mm}
\begin{IEEEbiography}
[{\includegraphics[width=1in,height=1.25in,clip,keepaspectratio]{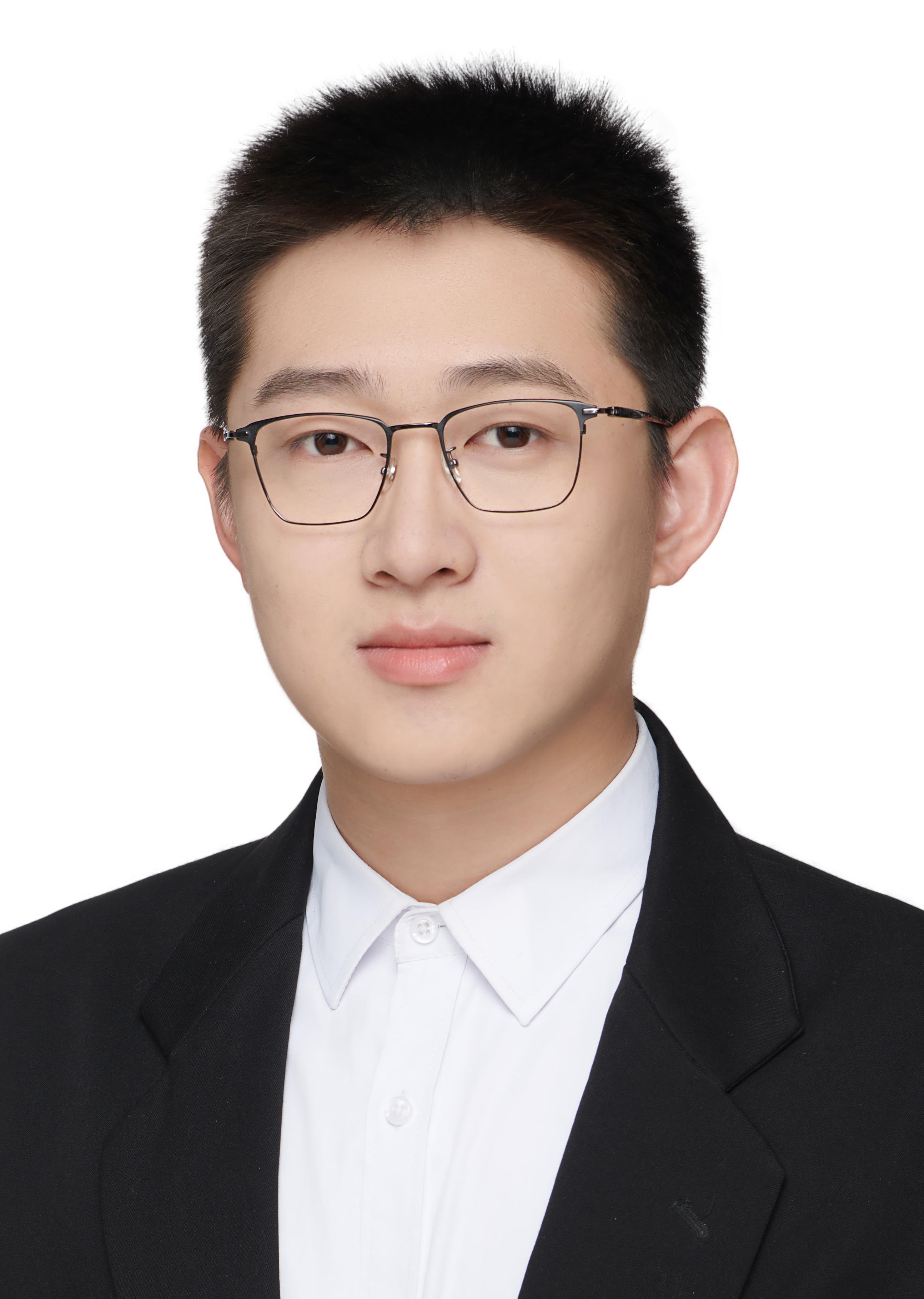}}]
{Kai Liu}
received the B.S. degree from Shanghai Jiao Tong University, Shanghai, China in 2024.
He is currently a first-year Ph.D. student at Shanghai Jiao Tong University.
His research interests include computer vision, model compression, and model quantization.
He is the co-organizer of NTIRE 2025 for real-world face restoration and image super-resolution ($\times$4).
He has published 2 papers in top-tier conferences on NeurIPS and ICML.
He serves as Reviewer for NeurIPS, ICCV, ICLR, CVPR, etc.
\end{IEEEbiography}

\vspace{ -15mm} 
 \begin{IEEEbiography}
[{\includegraphics[width=1in,height=1.25in,clip,keepaspectratio]{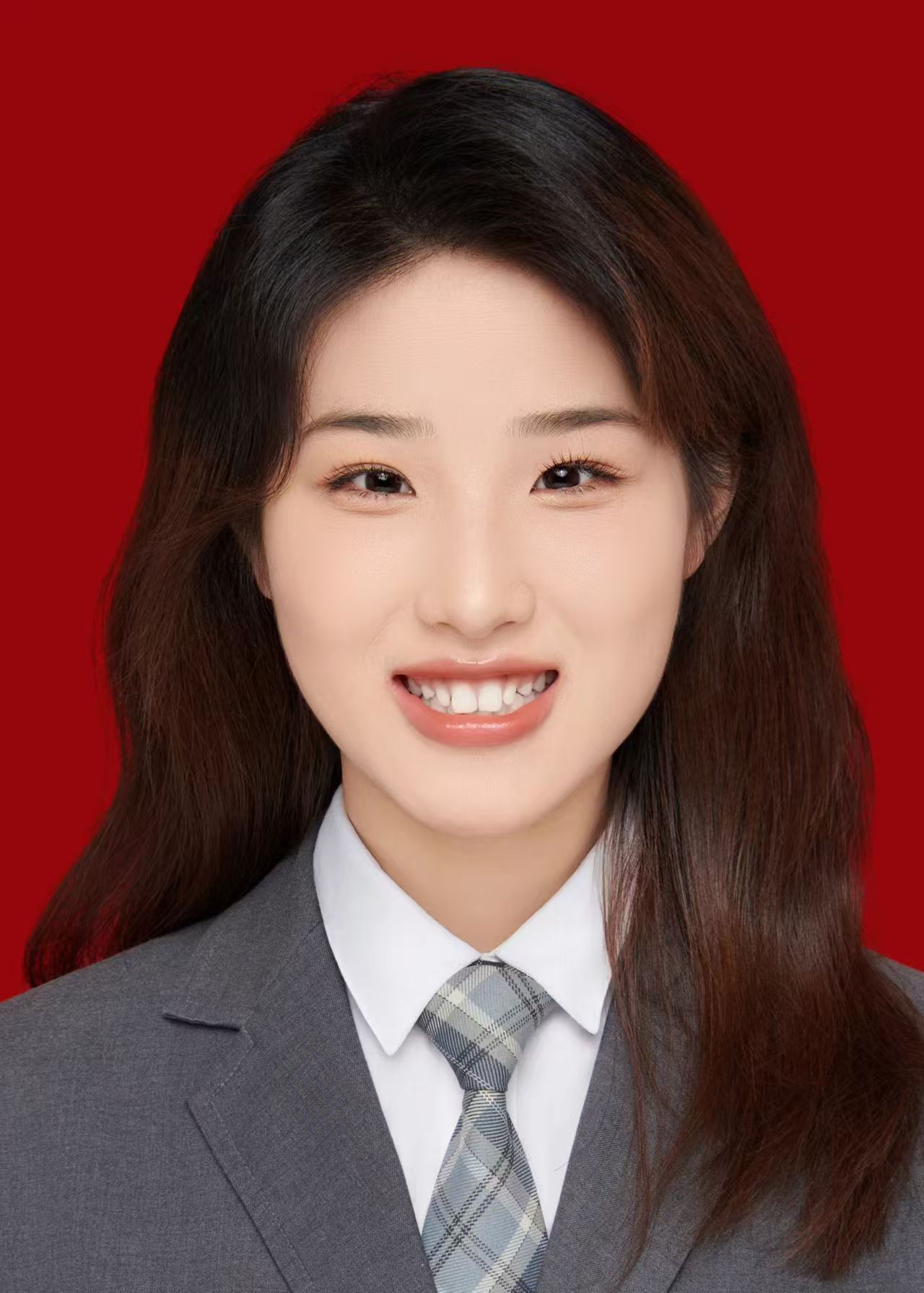}}]
{Qian Zheng}
is pursuing her bachelor's degree in the School of Integrated Circuit at Shanghai Jiao Tong University, Shanghai, China.
Her research interests lie in model compression.
\end{IEEEbiography}

\vspace{ -15mm}
\begin{IEEEbiography}
[{\includegraphics[width=1in,height=1.25in,clip,keepaspectratio]{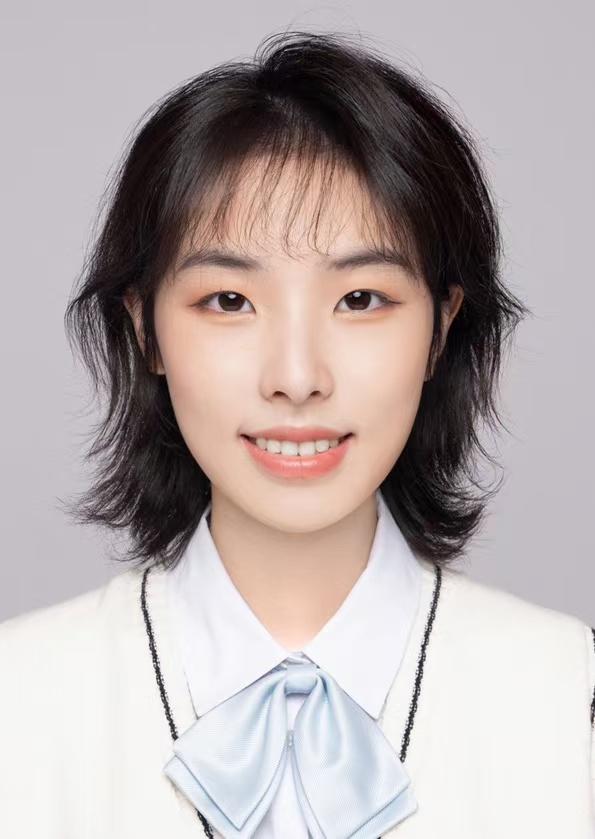}}]
{Kaiwen Tao}
is pursuing her bachelor's degree in the School of Computer Science at Shanghai Jiao Tong University, Shanghai, China.
Her research interests lie in model compression.
\end{IEEEbiography}

\vspace{ -15mm}
\begin{IEEEbiography}
[{\includegraphics[width=1in,height=1.25in,clip,keepaspectratio]{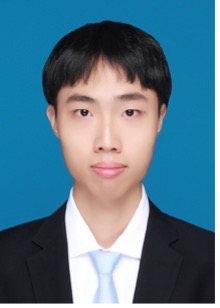}}]
{Zhiteng Li}
received the B.S. degree from Shanghai Jiao Tong University, Shanghai, China in 2023.
He is currently a second-year Master student at Shanghai Jiao Tong University.
His research interests include AIGC, model compression, and model quantization.
He has published 2 papers in top-tier conferences on ICLR.
He serves as Reviewer for NeurIPS, ICCV, ICLR, CVPR, etc.
\end{IEEEbiography}

\vspace{ -15mm}
\begin{IEEEbiography}[{\includegraphics[width=1in,height=1.25in,clip,keepaspectratio]{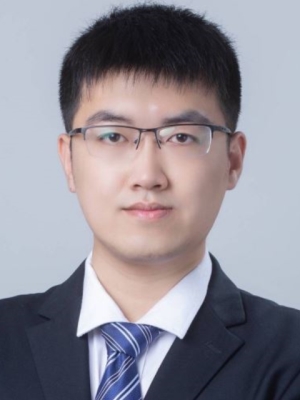}}]{Haotong Qin} is a Postdoctoral Researcher at the Center for Project-Based Learning (PBL) D-ITET at ETH Zürich. He was a Ph.D. student in the State Key Laboratory of Complex and Critical Software Environment at Beihang University. He obtained a B.Eng degree in computer science and engineering from Beihang University. His research interests include model compression and deployment toward efficient deep learning in real-world scenarios. He has published 35 papers in top-tier journals and conferences, such as IEEE TPAMI, IEEE TNNLS, IJCV, ICML, NeurIPS, ICLR, and CVPR. He serves as Guest Editor for Neural Networks, Area Chair in BMVC, and Reviewer for IEEE TPAMI, IEEE TIP, IEEE TNNLS, CVPR, etc.
\end{IEEEbiography}

\vspace{ -15mm}
\begin{IEEEbiography}[{\includegraphics[width=1in,height=1.25in,clip,keepaspectratio]{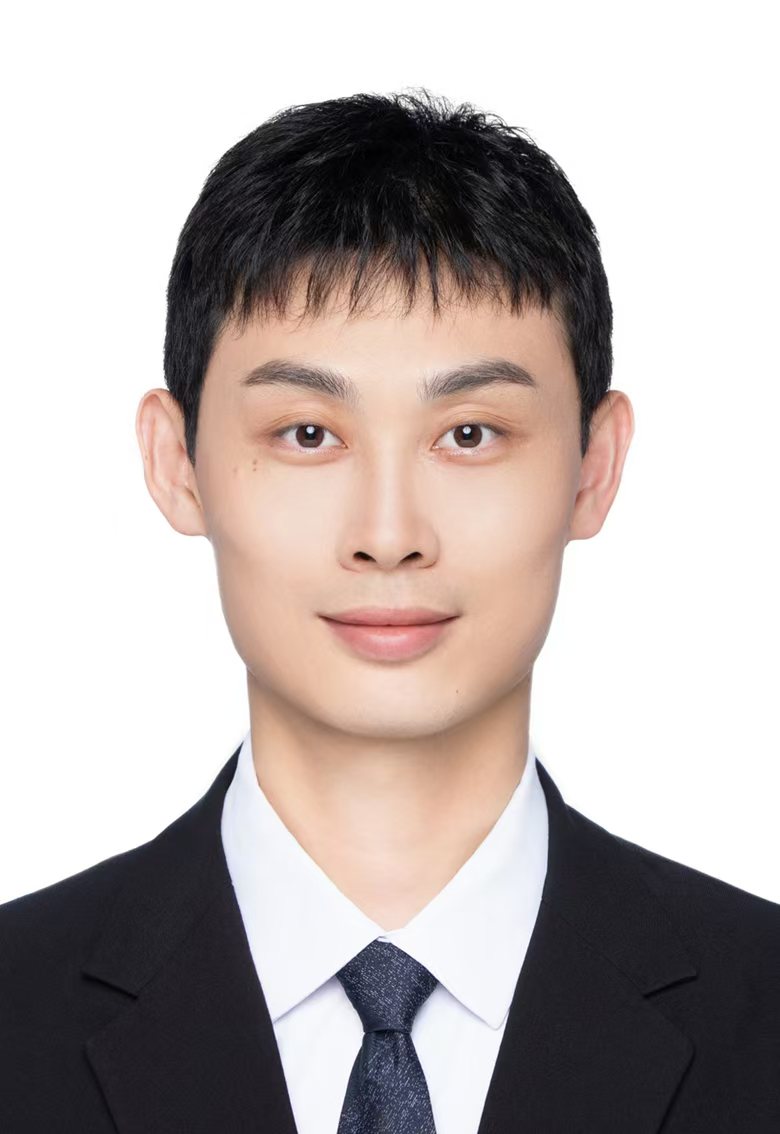}}]{Wenbo Li} received the BEng and MS degrees from Shanghai Jiao Tong University, in 2016 and 2019, and the PhD degree in the Department of Computer Science and Engineering from The Chinese University of Hong Kong, in 2023. He is currently a research scientist in Huawei Noah's Ark Lab. His primary research interests lie in low-level computer vision and AIGC. His paper was selected in the CVPR 2022 Best Paper Finalists. He also serves as a reviewer for TPAMI, IJCV, TIP, ICML, ICLR, NeurIPS, CVPR, etc.
\end{IEEEbiography}

\vspace{ -15mm}
\begin{IEEEbiography}[{\includegraphics[width=1in,height=1.25in,clip,keepaspectratio]{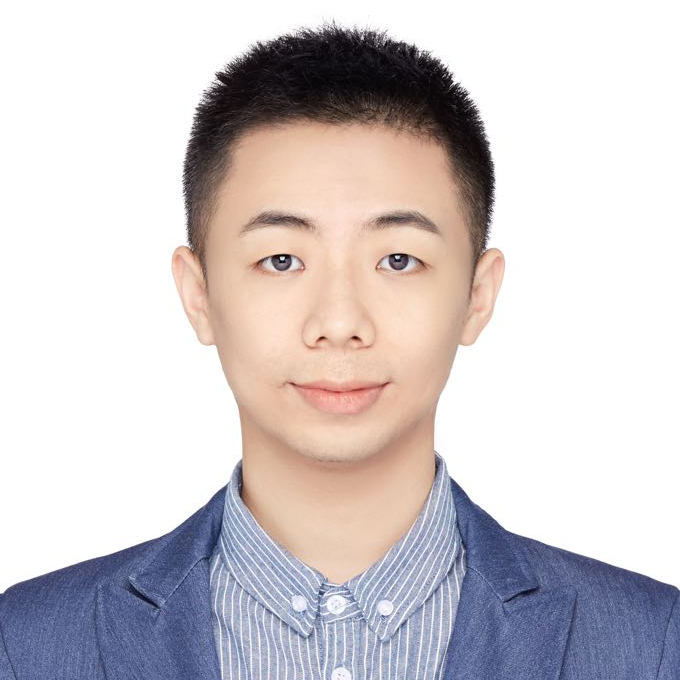}}]{Yong Guo} obtained his PhD degree with the School from South China University of Technology in 2021. After that, he worked as a post-doctoral researcher at the Computer Vision and Machine Learning Group in Max Planck Institute for Informatics (MPI-INF), supervised by Prof. Bernt Schiele. In 2023, he joined Huawei as a senior researcher. His research interests include model compression and image restoration.
\end{IEEEbiography}

\vspace{ -15mm}
\begin{IEEEbiography}[{\includegraphics[width=1in,height=1.25in,clip,keepaspectratio]{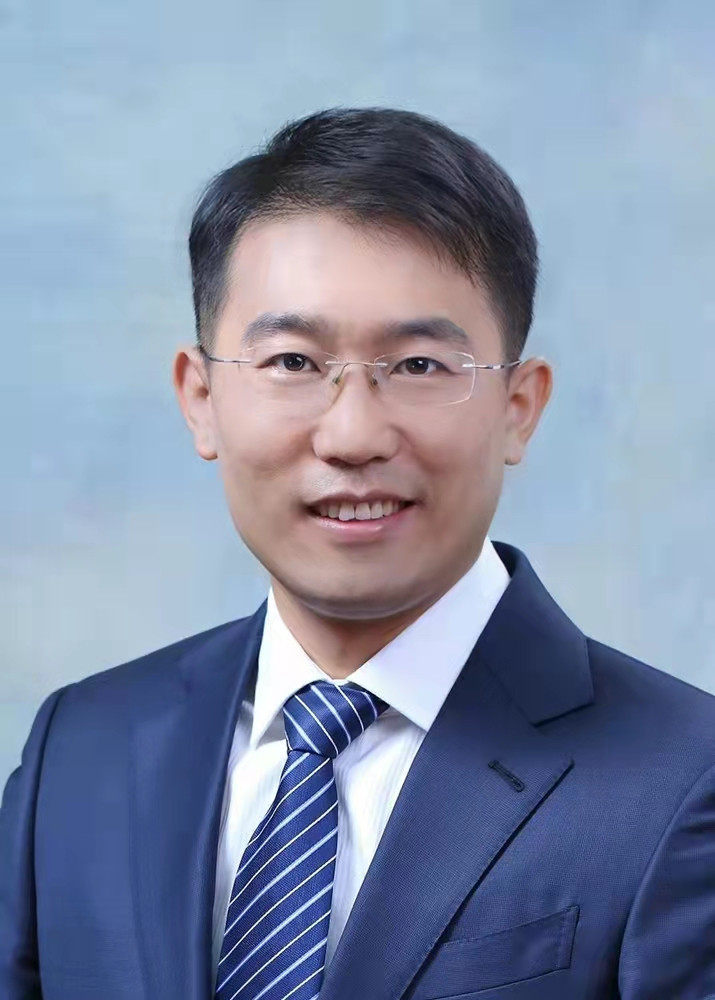}}]{Xianglong Liu} is currently a Professor, serves as the vice dean of the School of Computer Science and Engineering at Beihang University, and is also the deputy director of the State Key Laboratory of Complex and Critical Software Environment. He received his B.S. and Ph.D. degrees under the supervision of Prof. Wei Li and visited the DVMM Lab at Columbia University as a joint PhD student supervised by Prof. Shih-Fu Chang. He is the recipient of the China National Excellent Youth Science Fund. He has published over 100 papers in top conferences/journals in artificial intelligence and information security, such as NeurIPS, ICLR, CVPR, ICCV, CSS, and IJCV. He serves as Associate Editor and Guest for several SCI journals like Pattern Recognition and IET Image Processing and as Promotion Editor for journals like Frontiers of Computer Science and Acta Aeronautica et Astronautica Sinica. He serves as Area Chair in top conferences such as AAAI and ACM MM and has frequently organized workshops and competitions in conferences like CVPR, IJCAI, and AAAI.
\end{IEEEbiography}

\vspace{ -15mm}
\begin{IEEEbiography}[{\includegraphics[width=1in,height=1.25in,clip,keepaspectratio]{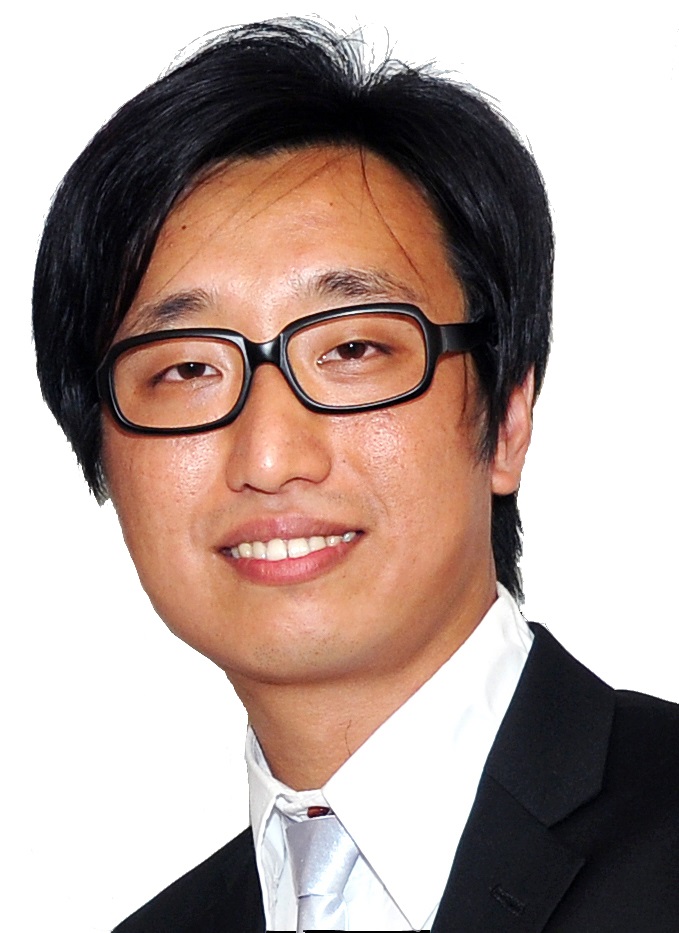}}]{Linghe Kong} is currently a professor with the Department of Computer Science and Engineering at Shanghai Jiao Tong University. He previously served as a postdoctoral researcher at Columbia University, McGill University, and Singapore University of Technology and Design. He received his Ph.D. degree at Shanghai Jiao Tong University, 2013, his Master degree at TELECOM SudParis (ex. INT), 2007, and his Bachelor degree at Xidian University, 2005. His research interests include Internet of things, satellite network, big data, and artificial intelligence.
\end{IEEEbiography}

\vspace{ -15mm}
\begin{IEEEbiography}[{\includegraphics[width=1in,height=1.25in,clip,keepaspectratio]{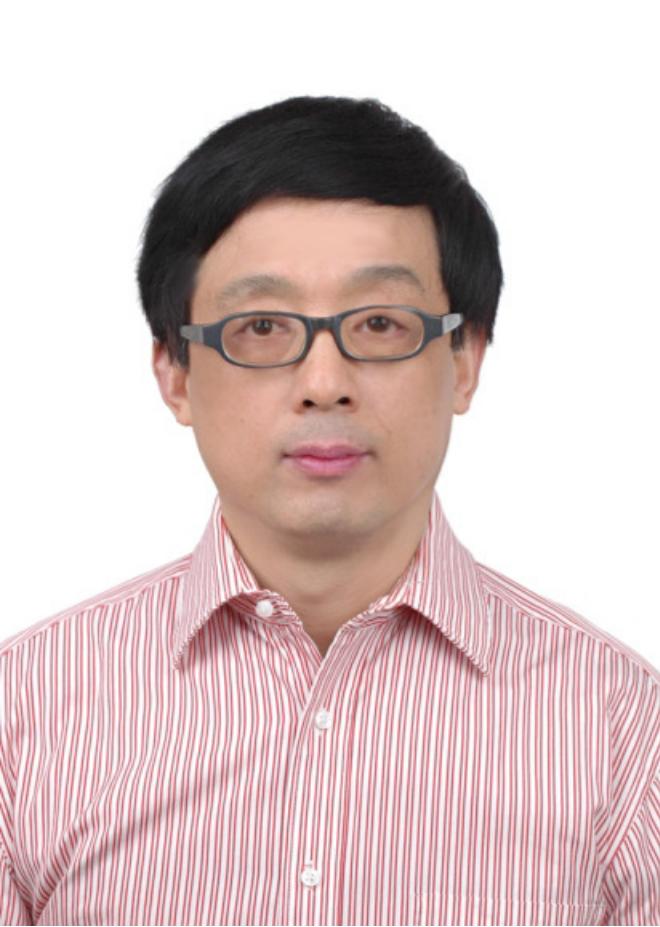}}]{Guihai Chen} (Fellow, IEEE) received the B.S. degree in computer software from Nanjing University, China, in 1984, the M.E. degree in computer applications from Southeast University in 1987, and the Ph.D. degree in computer science from The University of Hong Kong in 1997. He was a Visiting Professor with many universities, including the Kyushu Institute of Technology, Japan, in 1998, The University of Queensland, Australia, in 2000, and Wayne State University, USA, from September 2001 to August 2003. He is currently a Professor at Shanghai Jiao Tong University. He has a wide range of research interests with a focus on sensor networks, peer-to-peer computing, high-performance computer architecture, and combinatorics.
\end{IEEEbiography}

\vspace{ -15mm}
\begin{IEEEbiography}[{\includegraphics[width=1in,height=1.25in,clip,keepaspectratio]{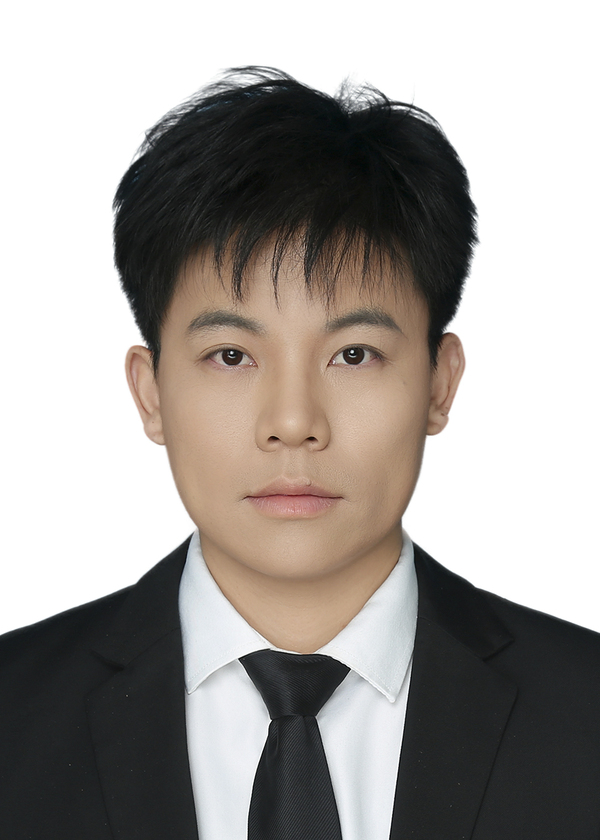}}]{Yulun Zhang} received a B.E. degree from the School of Electronic Engineering, Xidian University, China, in 2013, an M.E. degree from the Department of Automation, Tsinghua University, China, in 2017, and a Ph.D. degree from the Department of ECE, Northeastern University, USA, in 2021. He is an associate professor at Shanghai Jiao Tong University, Shanghai, China. He was a postdoctoral researcher at Computer Vision Lab, ETH Zürich, Switzerland. His research interests include image/video restoration and synthesis, biomedical image analysis, model compression, multimodal computing, large language model, and computational imaging. He is/was an Area Chair for CVPR, ICCV, ECCV, NeurIPS, ICML, ICLR, IJCAI, ACM MM, and a Senior Program Committee (SPC) member for IJCAI and AAAI.
\end{IEEEbiography}

\vspace{ -15mm}
\begin{IEEEbiography}[{\includegraphics[width=1in,height=1.25in,clip,keepaspectratio]{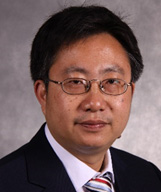}}]{Xiaokang Yang} (Fellow, IEEE) is currently a distinguished professor of School of Electronic Information and Electrical Engineering, and the deputy director of the Institute of Image Communication and Information Processing, Shanghai Jiao Tong University, Shanghai, China. He received the Ph.D. degree from Shanghai Jiao Tong University, Shanghai, China, in 2000. He has published over 200 refereed papers, and has filed 60 patents.
\end{IEEEbiography}



\section*{Supplementary material}

# Supplementary File: Low-bit Model Quantization for Deep Neural Networks: A Survey

Kai Liu†, Qian Zheng†, Kaiwen Tao†, Zhiteng Li, Haotong Qin, Wenbo Li, Yong Guo, Xianglong Liu, Linghe Kong*, Guihai Chen, *Fellow, IEEE*, Yulun Zhang*, Xiaokang Yang, *Fellow, IEEE*

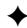

## 1 COMMON INFORMATION OF QUANTIZATION

Different tasks, such as image classification, image generation, and text understanding, usually need different quantization methods due to the varying characteristics of weights and activation. Here, we collect the frequently studied 7 tasks in the field of quantization, including text-to-text (T2T), image-to-text (I2T), visual generation, super-resolution, image segmentation, image classification, and object detection. Also, we collect the commonly used datasets and models in these tasks, and representative papers, as shown in Tab 1. We tried to provide the performance comparison of different quantization methods on these benchmarks, but failed. This is because the models, datasets, and quantification schemes adopted by the recent methods are all different, leading to an exponential number of experiments. We hope that Tab 1 can help the emerging forces in the model quantization quickly understand the cutting edge of quantization methods and thereby promote the development.

## 2 RELATED WORKS

Many works also investigate the model compression techniques [89]–[96].

Rokh *et al.* [90] provides a comprehensive overview of model quantization techniques for deep neural networks (DNNs) in the context of image classification. The authors delve into the fundamental concepts of quantization, exploring various methods and approaches that have been proposed to reduce the memory footprint and computational complexity of DNNs while maintaining their accuracy. The

- † denotes equal contribution.
- Kai Liu, Qian Zheng, Kaiwen Tao, Zhiteng Li, Linghe Kong, Guihai Chen, Yulun Zhang, and Xiaokang Yang are with the School of Computer Science, Shanghai Jiao Tong University, Shanghai, China. (Email: normal.kliu@gmail.com, xiaozheng2023@sjtu.edu.cn, miukiii19@sjtu.edu.cn, ieeezhitengli@gmail.com, linghe.kong@sjtu.edu.cn, gchen@cs.sjtu.edu.cn, yulun100@gmail.com, xkyang@sjtu.edu.cn.
- Haotong Qin is with ETH Zürich, Switzerland. (Email: qinhaotong@gmail.com).
- Wenbo Li is with Huawei Noah's Ark Lab, China. (Email: wenboli@cse.cuhk.edu.hk). Yong Guo is with Huawei Consumer Business Group, China. (E-mail: guoyongcs@gmail.com).
- Xianglong Liu is with Beihang University, China. (Email: xlliu@buaa.edu.cn).
- Yulun Zhang and Linghe Kong are the corresponding authors.

survey categorizes and discusses different quantization strategies, including clustering-based methods, uniform and non-uniform quantization, and the use of scale factors to enhance quantization accuracy. Additionally, it examines the training of quantized neural networks, highlighting the role of techniques such as the straight-through estimator and quantization regularization. The sensitivity of different layers to quantization is analyzed, and the performance of state-of-the-art methods on benchmark datasets like CIFAR-10 and ImageNet is presented. However, it fails to cover quantization methods in other areas, such as large language models and other computer vision tasks. On the contrary, we provide a more comprehensive and updated view of model quantization.

Gholami *et al.* [91] provides a comprehensive overview of quantization methods aimed at enhancing the efficiency of neural network inference. It delves into the fundamental concepts and advanced techniques employed in quantizing neural networks, highlighting the trade-offs between precision, accuracy, and computational efficiency. The authors categorize various quantization approaches, including uniform and non-uniform quantization, symmetric and asymmetric schemes, and different calibration methods. They also discuss the impact of quantization granularity, ranging from layer-wise to channel-wise quantization, and the implications for hardware deployment. The survey further explores the integration of quantization with other optimization techniques such as knowledge distillation and pruning, emphasizing the potential for mixed-precision quantization to balance accuracy and efficiency. Additionally, it examines the challenges and opportunities in hardware-aware quantization, particularly for edge devices.

The survey by Shen *et al.* [95] provides a comprehensive exploration of quantization techniques for LLMs, categorizing them into three primary approaches: post-training quantization, quantization-aware fine-tuning, and quantization-aware training. PTQ methods, which apply quantization after pre-training, offer simplicity and efficiency but may introduce performance degradation. QAF techniques integrate quantization during the fine-tuning phase to compensate for some of the quantization losses, achieving a balance between accuracy and computational efficiency. QAT, which incorporates quantization directly into the training process, enables models to adapt to low-precision representations

TABLE 1
Common Datasets, Models, and Representative Papers for Various Tasks

| Tasks | Training Dataset | Testing Dataset | Models | Representative Papers |
|---|---|---|---|---|
| T2T | Wikitext [1], WikiText2 [1], C4 [2], HH-RLHF [3], Alpaca [4], Flan v2 [5], SQuAD1.1 [6] | Wikitext [1], C4 [2], WikiText2 [1], PTB [7], GLUE [8], Super-NaturalInstructions [9], PIQA [10], ARC-e, ARC-c, HellaSwag [11], Winogrande [12], | Transformer [13], OPT [14], Bloom [15], GLM [16], MT-NLG [17], Llama-2 [18], Falcon [19], Mixral [20], BERT [21], RoBERTa [22], XLNet [23], GPT-3 [24], GPT-4 [25], T5 [2], Mistral [26], Starcoder [27], Gemma [28], [29] | ZeroQuant [30], Smoothquant [31], Squeezellm [32], Q-BERT [33], Qlora [34], GPTQ [35], SpQR [36], AWQ [37], [38] |
| I2T | Obelics [39], LAION-5B [40], Wikipedia-based Image Text [41] | GQA [42], TextVQA [43], ScienceQA [44], VizWiz [45] | VICUNA-V1.5 [], LLAVA-V1.5 [46], MoE-LLaVA, OpenFlamingo [47] | [48] |
| Visual Generation | ImageNet [49], CIFAR-10 [50], LSUN, COCO [51] | ImageNet [49], CIFAR-10 [50], LSUN-Bedrooms [52], LSUN-Churches [52], COCO [51], SQuAD1.1 [6], VQA-v2 [53], GQA [42], VizWiz [45], SQAI [44], VQAT [43] | DDIM [54], DDPM [55], ResNet [56], MobileNetV2 [57], Swin , BERT [21], RoBERTa [22], LLaVA [46], FFHQ [58], VILA [59], transformer [60] | AWQ [37] QDROP [61], PTQ4DM [62], [63], [64], |
| Super Resolution | DIV2K [65], DRealSR [66], | Set5 [67], Set14 [68], B100 [69], Urban100 [70], DRealSR [66] | VDSR [71], EDSR [72], RDN [73], SRRes-Net [74] | DAQ [75] SLB [76], [77], |
| Image Segmentation | COCO [51], ImageNet [49] | ImageNet [49] | Enet, DeepLabV3+ [78], RepVGG [79] | AdaRound [80] |
| Image Classification | ImageNet [49], CIFAR-10 [50], ILSVRC2012 | ImageNet [49], CIFAR-10 [50], ILSVRC2012 | Enet, ResNet [56], MobileNetV2 [57], RegNet [81], MnasNet [82], RepVGG [79], VGG-Small [83], Inception-V3 [84] | QDROP [61], BRECQ [85], HAWQ [86], [87] |
| Object Detection | COCO [51] | COCO [51] | One-stage RetinaNet [88], ResNet [56], Enet, RepVGG [79] | QDROP [61], BRECQ [85] |

from the outset, potentially yielding more robust and efficient quantized models. The survey also highlights experimental studies that uncover underlying patterns and challenges in LLM quantization, providing valuable insights for future research. However, this survey only investigates a limited papers and only focuses on LLM instead of other advanced models.

As LLMs become crucial, some works thoroughly evaluate quantization methods on LLMs [92], [94], [96]. Li *et al.* [92] present a thorough evaluation of efficiency and accuracy by evaluating the effect of PTQ on weight, activation, and KV Cache on 11 model families. While Dettmers *et al.* [96] study the trade-off between the size of the base model and bit-width by developing inference scaling laws of zero-shot performance in LLMs to determine the bit-precision and model size that maximize performance. Yao *et al.* [94] conduct a comprehensive experiments of diverse PTQ on weight-only, activation-only, and weight-and-activation quantization.

## 3 Other Model Compression Techniques

**a) Pruning:** Pruning aims to remove redundant structures to compress DNNs. Redundancies is identified based on different criteria such as sparsity [97]–[99], Bayesian pruning [100], [101], importance ranking [102], [103], grouped kernel search [104], reinforcement learning [105], and the

lottery ticket hypothesis [106]. Pruning strategies are broadly classified into two categories: structured and unstructured. Structured pruning [107], [108] involves the removal of entire neurons or filters, while unstructured pruning [109] operates at the weight level by removing individual weights or neurons based on their importance. Structure pruning enjoys better alignment with existing hardware and software acceleration frameworks, but suffers from bigger performance degradation. As for unstructured pruning, finer granularity allows a sparser model and higher precision in preserving model performance.

**b) Knowledge Distillation:** Knowledge distillation [110]–[113], also known as model distillation, is a technique for transferring knowledge from the teacher model to the student model, while the former is large and complex and the latter is small and simple. The core idea behind this method is that a well-trained teacher model encodes explicit knowledge in its weights and also implicitly captures deep-level features and patterns in its internal representations, which can be passed on to the student model through specific technical means. In the current context, knowledge distillation has been developed into many variants, and one-step distillation for diffusion models is increasingly popular.

**c) Model Design:** Lightweight network design focuses on optimizing the model architecture to reduce the number of parameters. In recent years, various lightweight network

architectures [60], [114], [115] have been proposed. These architectures incorporate modern designs, such as attention mechanisms and Mamba architecture, to reduce model complexity while maintaining high performance.

## 4 Why Quantization Brings Acceleration

The acceleration of quantization is mainly attributed to the following reasons:

*Memory Access Acceleration.* Quantization alleviates memory bottlenecks through precision reduction, compressing weight and activation representations. Since modern computing chips often suffer from memory bottlenecks where computational power far exceeds memory throughput, quantization directly reduces the memory footprint and the volume of data transfer. Additionally, theoretical analysis has consistently demonstrated that memory access optimization often dominates overall speedup despite computational overhead from weight dequantization.

*Vectorization.* Efficient vectorized computing instructions are fundamentally engineered to exploit data-level parallelism in large-scale computational workloads. For example, a 512-bit register natively processes 16 operands of FP32. With INT8 quantization, equivalent hardware resources enable 64-element parallel execution. Moreover, integer operations often take less time and energy than floating-point operations. Compiler-assisted optimizations further maximize hardware utilization through instruction scheduling and memory access pattern alignment.

Researchers also focus on the overhead brought by the quantization and dequantization processes. Techniques like operator fusion and computation graph optimization are investigated to minimize or eliminate the overhead.



\end{document}